\documentclass[conference]{IEEEtran}
\IEEEoverridecommandlockouts
\usepackage{cite}
\usepackage{amsmath,amssymb,amsfonts}
\usepackage{algorithmic}
\usepackage{graphicx}
\usepackage{textcomp}
\usepackage{xcolor}
\usepackage{svg}
\usepackage{enumitem}
\usepackage{caption}
\usepackage{subfig}
\usepackage{hyperref}
\def\BibTeX{{\rm B\kern-.05em{\sc i\kern-.025em b}\kern-.08em
T\kern-.1667em\lower.7ex\hbox{E}\kern-.125emX}}
\begin{document}
\title{Natural Emergence of Heterogeneous Strategies\\in Artificially Intelligent Competitive Teams
\thanks{\noindent The authors are with the Robotics Institute, Carnegie Mellon University, Pittsburgh, PA 15213, USA. Email: \texttt{\{adeka, katia\}@cs.cmu.edu}.
Supplementary video: \href{https://youtu.be/ltHgKYc0F-E}{\texttt{https://youtu.be/ltHgKYc0F-E}}
Code: {\scriptsize \href{https://github.com/Ankur-Deka/Emergent-Multiagent-Strategies}{ \texttt{http://github.com/Ankur-Deka/Emergent-Multiagent-Strategies}}}
This work was funded by DARPA award \#HR00111920028.}}

\author{\IEEEauthorblockN{Ankur Deka and Katia Sycara}
\vspace{-1cm}
}

\maketitle
\begin{abstract}
Multi agent strategies in mixed cooperative-competitive environments can be hard to craft by hand because each agent needs to coordinate with its teammates while competing with its opponents. Learning based algorithms are appealing but many scenarios require heterogeneous agent behavior for the team's success and this increases the complexity of the learning algorithm. In this work, we develop a competitive multi agent environment called FortAttack in which two teams compete against each other. We corroborate that modeling agents with Graph Neural Networks and training them with Reinforcement Learning leads to the evolution of increasingly complex strategies for each team. We observe a natural emergence of heterogeneous behavior amongst homogeneous agents when such behavior can lead to the team's success. Such heterogeneous behavior from homogeneous agents is appealing because any agent can replace the role of another agent at test time. Finally, we propose ensemble training, in which we utilize the evolved opponent strategies to train a single policy for friendly agents.
\end{abstract}
\vspace{2mm}

\section{Introduction}

Multi agent systems can play an important role in scenarios such as disaster relief, defense against enemies and games. There have been studies on various aspects of it including task assignment, \cite{strata}, \cite{shishika2020cooperative}, resilience to failure, \cite{ramachandran2019resilience}, scalability \cite{agarwal2019learning} and opponent modeling, \cite{wen2019probabilistic}. Multi agent systems become increasingly complex in mixed cooperative-competitive scenarios where an agent has to cooperate with other agents of the same team to jointly compete against the opposing team. It becomes difficult to model behavior of an agent or a team by hand and learning based methods are of particular appeal.


Our goal is to develop a learning based algorithm for decentralized control of multi agent systems in mixed cooperative-competitive scenarios with the ability to handle a variable number of agents, as some robots may get damaged in a real world scenario or some agents may get killed in a game. To be able to handle a variable number of agents and to scale to many agents, we propose to use a graph neural network (GNN) based architecture to model inter-agent interactions, similar to \cite{agarwal2019learning} and \cite{baker2019emergent}. This approach relies on shared parameters amongst all agents in a team which renders all of them homogeneous. We aim to study if heterogeneous behavior can emerge out of such homogeneous agents.\\

    
    
    

Our contributions in this work are:
\begin{itemize}
    \item We have developed a mixed cooperative-competitive multi agent environment called FortAttack with simple rules yet room for complex multi agent behavior.
    \item We corroborate that using GNNs with a standard off the shelf reinforcement learning algorithm can effectively model inter agent interactions in a competitive multi agent setting.
    \item To train strong agents we need competitive opponents. Using an approach inspired by self play, we are able to create an auto curriculum that generates strong agents from scratch without using any expert knowledge. Strategies naturally evolved as a winning strategy from one team created pressure for the other team to be more competitive. We were able to achieve this by training on a commodity laptop.
    \item We show that highly competitive heterogeneous behavior can naturally emerge amongst homogeneous agents with symmetric reward structure when such behavior can lead to the team's success. Such behavior implicitly includes heterogeneous task allocation and complex coordination within a team, none of which had to be explicitly crafted but can be extremely beneficial for multi agent systems.
\end{itemize}

\section{Related work}

\begin{figure*}[ht]
    \centering
    \includegraphics[width=0.35\textwidth]{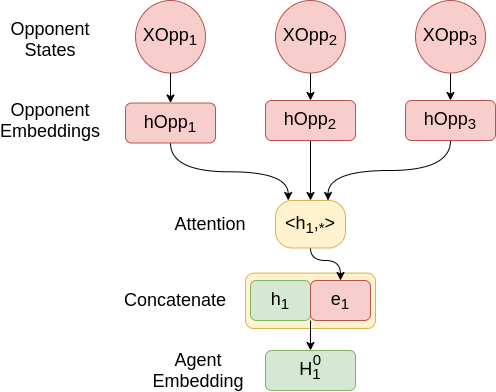}
    \quad \quad
    \includegraphics[width=0.23\textwidth]{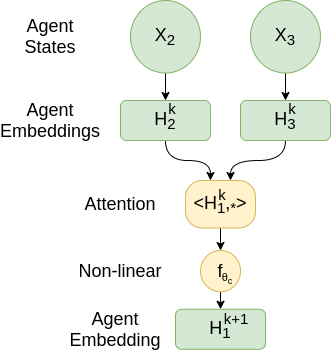}
    \caption{Modeling of inter agent interactions as a graph from the perspective of agent 1, in a 3 friendly agents vs 3 opponents scenario. Left: agent 1's embedding, $H^0_1$ is formed by taking into consideration the states of all opponents through an attention layer. Right: agent 1's embedding gets updated, $(H^k_1 \rightarrow H^{k+1}_1)$  by taking into consideration its team mates through an attention layer.}
    \label{fig:graphical_model}
    \vspace{-5mm}
\end{figure*}

The recent successes of reinforcement learning in games, \cite{mnih2015human}, \cite{silver2017mastering} and robotics, \cite{schulman2017proximal}, \cite{haarnoja2018soft} have encouraged researchers to extend reinforcement learning to multi agent settings. 

There are three broad categories of approaches used, centralized, decentralized and a mix of the two. Centralized approaches have a single reinforcement learning agent for the entire team, which has global state information and selects joint actions for the team. However, the joint state and action spaces grows exponentially with the number of agents rendering centralized approaches difficult to scale, \cite{bu2008comprehensive}.

Independent Q-learning, \cite{tan1993multi}, \cite{tampuu2017multiagent} is a decentralized approach where each agent learns separately with Q-learning, \cite{watkins1992q}  and treats all other agents as parts of the environment. Inter agent interactions are not explicitly modeled and performance is generally sub-par. 

Centralized learning with decentralized execution has gained attention because it is reasonable to remove communication restrictions at training time. Some approaches use a decentralized actor with a centralized critic, which is accessible only at training time. MADDPG, \cite{lowe2017multi} learns a centralized critic for each agent and trains policies using DDPG, \cite{lillicrap2015continuous}. QMIX, \cite{rashid2018qmix} proposes a monotonic decomposition of action value function. However, the use of centralized critic requires that the number of agents be fixed in the environment.

GridNet, \cite{han2019grid} addresses the issue of multiple and variable number of agents without exponentially growing the policy representation by representing a policy with an encoder-decoder architecture with convolution layers. However, the centralized execution realm renders it infeasible in many scenarios.

Graphs can naturally model multi agent systems with each node representing an agent. \cite{sukhbaatar2016learning} modeled inter agent interactions in multi agent teams using GNNs which can be learnt through back propagation. \cite{hoshen2017vain} proposed to use attention and \cite{agarwal2019learning} proposed to use an entity graph for augmenting environment information. However, these settings don't involve two opposing multi agent teams that both evolve by learning.

\cite{baker2019emergent} explored multi agent reinforcement learning for the game of hide and seek. They find that increasingly complex behavior emerge out of simple rules of the game over many episodes of interactions. However, they relied on extremely heavy computations spanning over many millions of episodes of environment exploration.

We draw inspiration from \cite{agarwal2019learning} and \cite{baker2019emergent}. For each team we propose to have two components within the graph, one to model the observations of the opponents and one to model the interactions with fellow team mates. Our work falls in the paradigm of centralized training with decentralized execution. We were able to train our agents in the FortAttack environment using the proposed approach on a commodity laptop. We believe that the reasonable computational requirement would encourage further research in the field of mixed cooperative-competitive MARL.


\section{Method}
We describe our method from the perspective of one team and use $X_i$ to denote the state of $i^{th}$ friendly agent in the team, which in our case is its position, orientation and velocity. We use $XOpp_{j}$ to denote the state of the $j^{th}$ opponent in the opposing team. Let $S = \{1,2,\dots,N_1\}$ denote the set of friendly agents and $S_{Opp} = \{N_1+1,N_1+2,\dots,N_1+N_2\}$ denote the set of opponents. Note that a symmetric view can be presented from the perspective of the other team. 

In the following, we describe how agent 1 processes the observations of its opponents and how it interacts with its teammates. Fig. \ref{fig:graphical_model} shows this pictorially for a 3 agents vs 3 agents scenario. All the other agents have a symmetric representation of interactions. 

\subsection{Modeling observation of opponents}
Friendly agent 1 takes its state, $X_1$ and passes it through a non-linear function, $f_{\theta_a}$ to generate an embedding, $h_1$. Similarly, it forms an embedding, $hOpp_j$ from each of its opponents with the function $f_{\theta_b}$.  
\begin{align}
    h_1 &= f_{\theta_a}(X_1)\\
    hOpp_j &= f_{\theta_b}(XOpp_j) \quad \forall j \in S_{Opp}
\end{align}

Note that the opponents don't share their information with the friendly agent 1. Friendly agent 1 merely makes its own observation of the opponents. It then computes a dot product attention, $\psi_{1j}$ which describes how much attention it pays to each of its opponents. The dimension of $h_1$ and $hOpp_j$ are $d_1$ each. This attention allows agent 1 to compute a joint embedding, $e_1$ of all of its opponents.

\begin{align}
    \hat{\psi}_{1j} &= \frac{1}{d_1}<h_1,hOpp_j>  \quad \forall j \in S_{Opp} \label{eq:dot_prod}\\
    \psi_{1j} &= \frac{\exp(\hat{\psi}_{1j})}{\sum_{m \in S_{Opp}} \exp(\hat{\psi}_{1m})}\\
    e_1 &= \sum_{j \in S_{Opp}} \psi_{1j}hOpp_j 
\end{align}

In Eq. \ref{eq:dot_prod}, $<,>$ denotes vector dot product. Note that $\sum_{j \in S_{Opp}}\psi_{1j} = 1$ which ensures that the net attention paid by agent 1 to its opponents is fixed. Finally, $e_1$ is concatenated with $h_1$ to form an agent embedding, $H_1^0$:
\begin{align}
    H_1^0 &= \text{concatenate}(h_1, e_1)
\end{align}

\subsection{Modeling interactions with teammates}
Agent 1 forms an embedding for each of its team mates with the non-linear function, $f_{\theta_a}$.
\begin{align}
    H_i^0 &= f_{\theta_a}(X_i) \quad \forall i \in S, i \neq 1
\end{align}
Dimension of $H_i^k, \forall i \in S$ is $d_2$. Agent 1 computes a dot product attention, $\phi_{1i}$ with all of its team mates and updates it's embedding with a non-linear function, $f_{\theta_c}$.
\begin{align}
    \hat{\phi}_{1i} &= \frac{1}{d_2}<H_1^k,H_i^k>  \quad \forall i \in S, i \neq 1 \label{eq:team_update_start}\\
    \phi_{1i} &= \frac{\exp(\hat{\phi}_{1i})}{\sum_{m \in S,m \neq 1} \exp(\hat{\phi}_{1m})}\\
    \hat{H}_1^{k+1} &= \sum_{i \in S, i \neq 1} \phi_{1i}H_i^k\\
    H_1^{k+1} &= f_{\theta_c}(\hat{H}_1^{k+1}) \label{eq:team_update_end}
\end{align}
Equations, \ref{eq:team_update_start} to \ref{eq:team_update_end} can be run over multiple iterations for $k=\{0,1,\dots,K\}$ to allow information propagation to other agents if agents can perceive only its local neighborhood similar to  \cite{agarwal2019learning}.

\subsection{Policy}
The final embedding of friendly agent 1, $H_1^K$ is passed through a policy head. In our experiments, we use a stochastic policy in discrete action space and hence the policy head has a sigmoid activation which outputs a categorical distribution specifying the probability of each action, $\alpha_m$.
\begin{align}
    \pi(\alpha_m|O_1) = \pi'(\alpha_m|H_1^K) = \text{sigmoid}(f_{\theta_d}(H_1^K))\\
    \text{where, } O_1 = \{X_i: i \in S\}\cup \{XOpp_j: j \in S_{Opp}\} \nonumber
\end{align}
Here, $O_1$ is the observation of agent 1, which consists of its own state and the states of all other agents that it observes. This corresponds to a fully connected graph. We do this for simplicity. In practice, we could limit the observation space of an agent within a fixed neighborhood around the agent similar to \cite{agarwal2019learning} and \cite{baker2019emergent}. 

\subsection{Scalability and real world applicability}
The learn-able parameters for a team are the shared parameters, $\theta_a, \theta_b, \theta_c$ and $\theta_d$ of the functions, $f_{\theta_a}, f_{\theta_b}, f_{\theta_c}$ and $f_{\theta_d}$, respectively which we model with fully connected neural networks. Note that the number of learn-able parameters is independent of the number of agents and hence can scale to a large number of agents. This also allows us to handle a varying number of agents as agents might get killed during an episode and makes our approach applicable to real world scenarios where a robot may get damaged during a mission.

\subsection{Training}
Our approach follows the paradigm of centralized training with decentralized execution. During training, a single set of parameters are shared amongst teammates. We train our multi agent teams with Proximal Policy Optimization (PPO), \cite{schulman2017proximal}. At every training step, a fixed number of interactions are collected from the environment using the current policy for each agent and then each team is trained separately using PPO.  

The shared parameters naturally share experiences amongst teammates and allow for training with fewer number of episodes. At test time, each agent maintains a copy of the parameters and can operate in decentralized fashion. We trained our agents on a commodity laptop with i7 processor and GTX 1060 graphics card.  Training took about 1-2 days without parallelizing the environment.

\section{Environment}

\begin{figure}
    \centering
    \includegraphics[width=0.6\columnwidth]{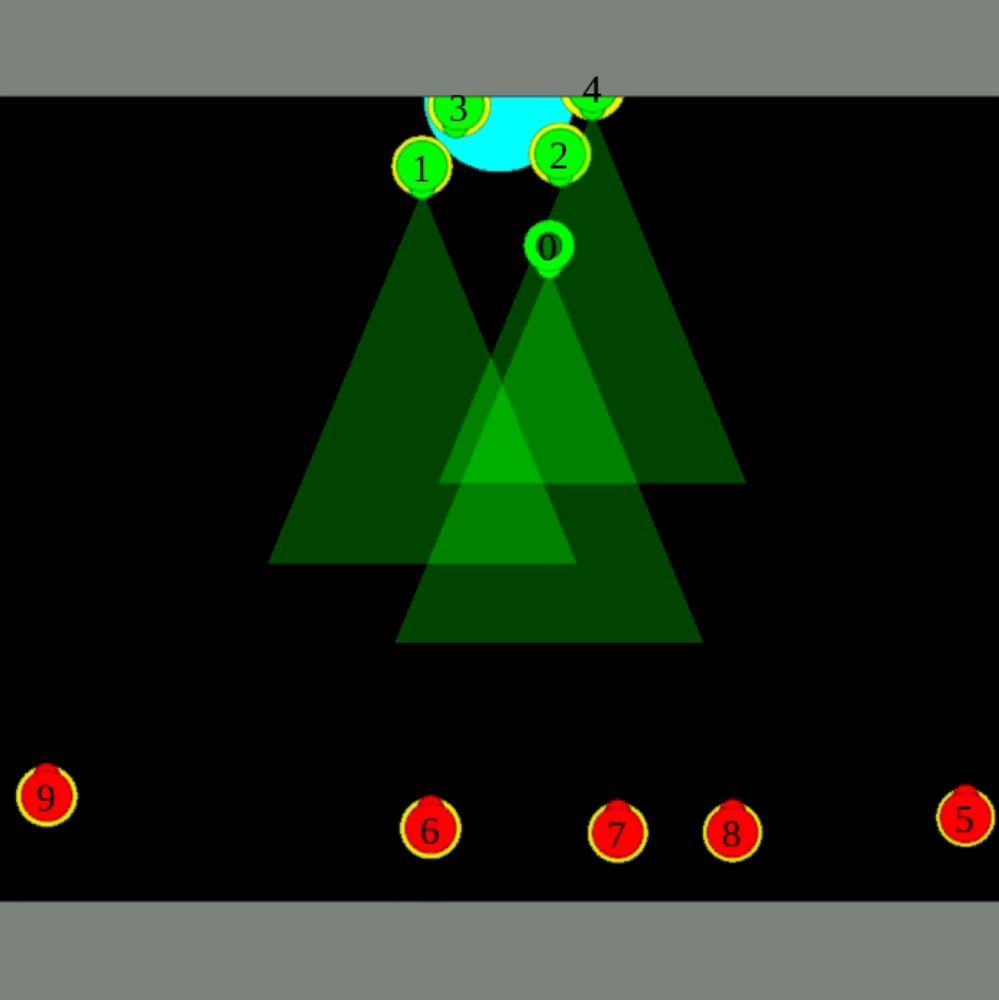}
    \caption{The Fortattack environment in which guards (green) need to protect the fort (cyan semicircle at the top) from the attackers (red). The attackers win when any one of them reaches the fort. Each agent can shoot a laser which can kill an opponent.}
    \label{fig:fortattack_env}
    \vspace{-5mm}
\end{figure}

We design a mixed cooperative-competitive environment called Fortattack with OpenAI Gym, \cite{brockman2016openai} like interface. Fig. \ref{fig:fortattack_env} shows a rendering of our environment. The environment consists of a team of guards, shown in green and a team of attackers, shown in red, that compete against each other. The attackers need to reach the fort which is shown as a cyan semi-circle at the top. Each agent can shoot a laser beam which can kill an opponent if it is within the beam window.

At the beginning of an episode, the guards are located randomly near the fort and the attackers are spawned at random locations near the bottom of the environment. The guards win if they manage to kill all attackers or manage to keep them away for a fixed time interval which is the episode length. The guards lose if even one attacker manages to reach the fort. The environment is built off of Multi-Agent Particle Environment, \cite{lowe2017multi}.

\begin{figure*}[ht]
    \centering
    \includegraphics[width=0.76\textwidth]{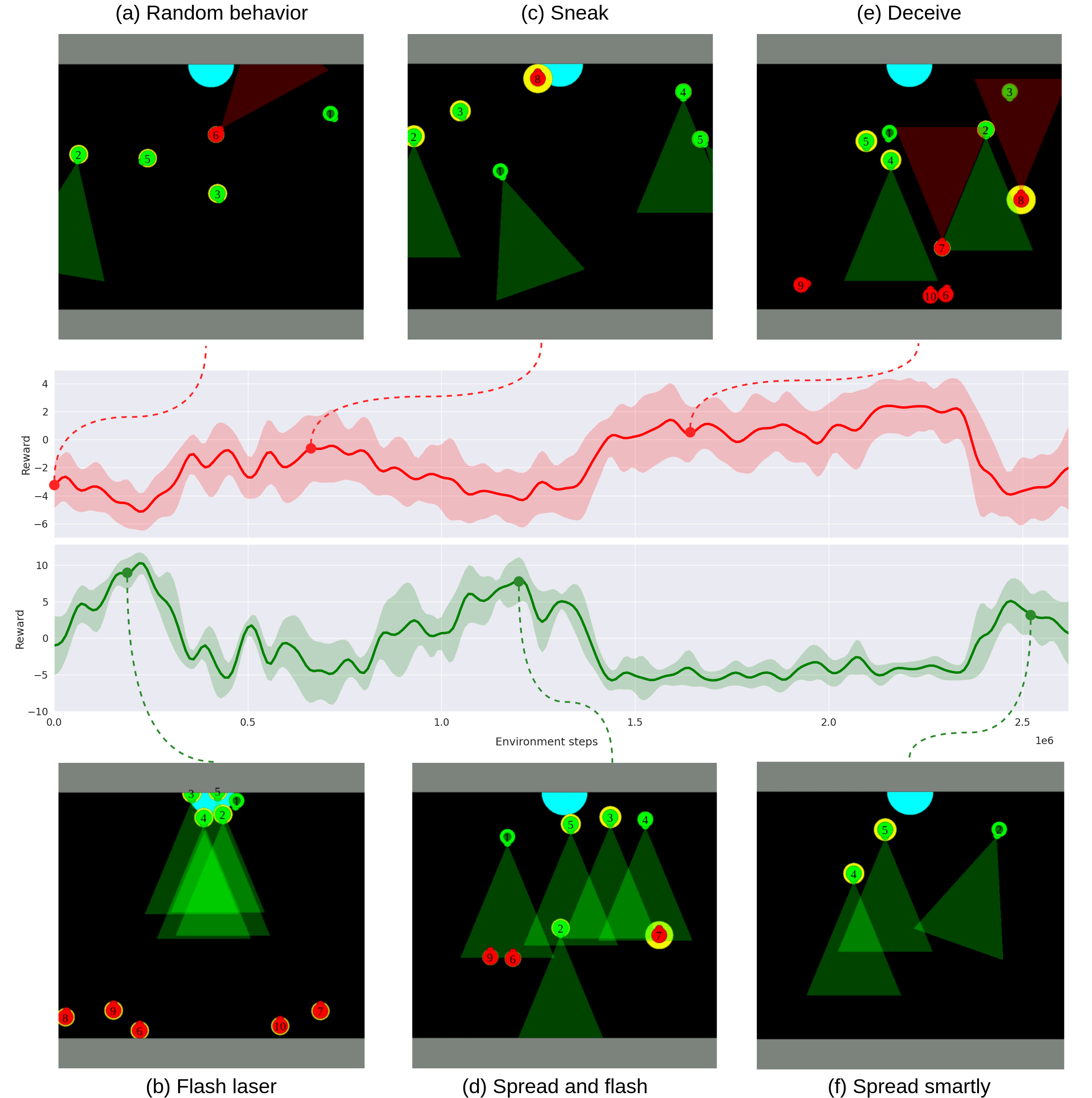}
    \caption{Average reward per agent per episode for the teams of attackers and guards as training progresses. The reward plots have distinct extrema and corresponding snapshots of the environment are shown. The x-axis shows the number of steps of environment interaction. The reward is plotted after Gaussian smoothing.}
    \label{fig:evolution}
    \vspace{-3mm}
\end{figure*}

\subsection{Observation space}
Each agent can observe all the other agents in the environment. Hence, the observation space consists of states (positions, orientations and velocities) of team mates and opponents. We assume full observability as the environment is small in size. This can possibly be extended to observability in the local neighborhood such as in \cite{agarwal2019learning} and \cite{baker2019emergent}.

\subsection{Action space}
At each time step, an agent can choose one of 7 actions, accelerate in $\pm x$ direction, accelerate in $\pm y$ direction, rotate clockwise/anti-clockwise by a fixed angle or do nothing.

\subsection{Reward structure}

Each agent gets a reward which has components of its individual and the team's performance. The rewards structure is described in more detail in the Appendix.



\begin{figure*}
    \centering
    \subfloat[Random exploration]
    {
        \centering
        \includegraphics[width=0.17\textwidth]{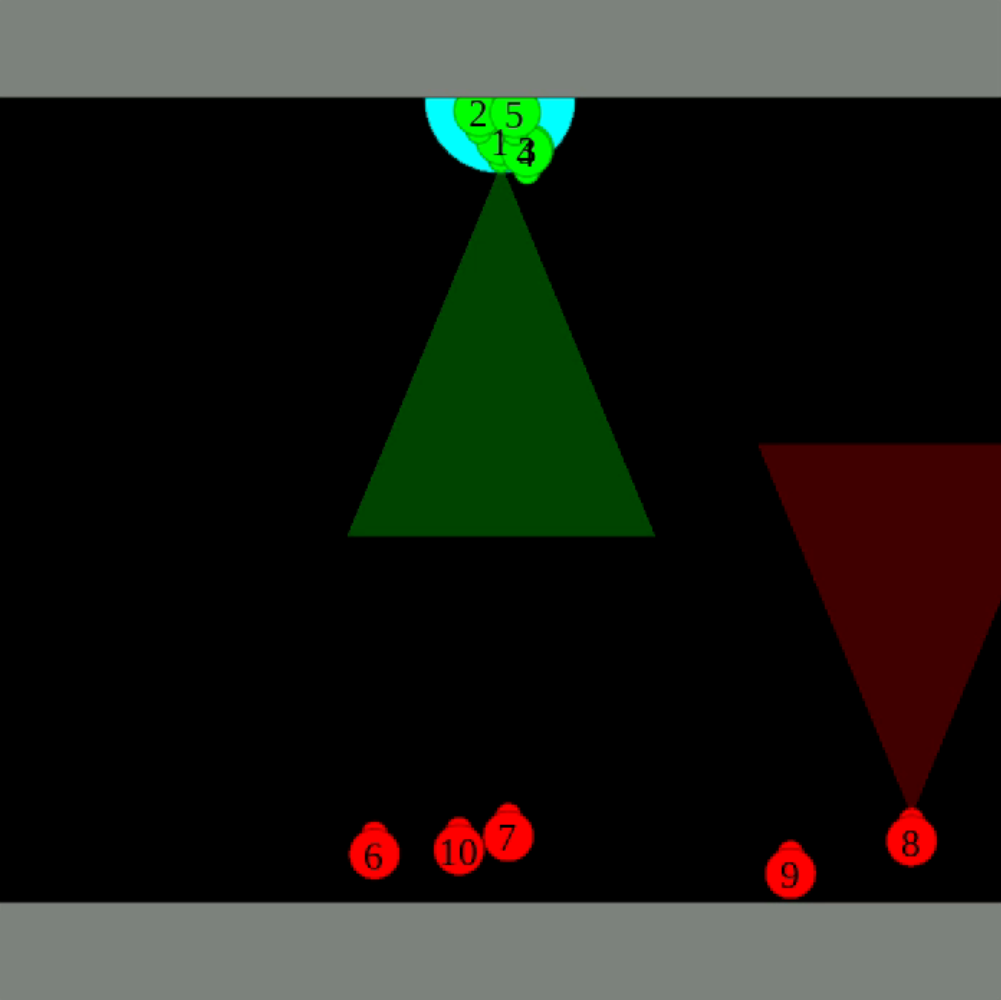}
        \includegraphics[width=0.17\textwidth]{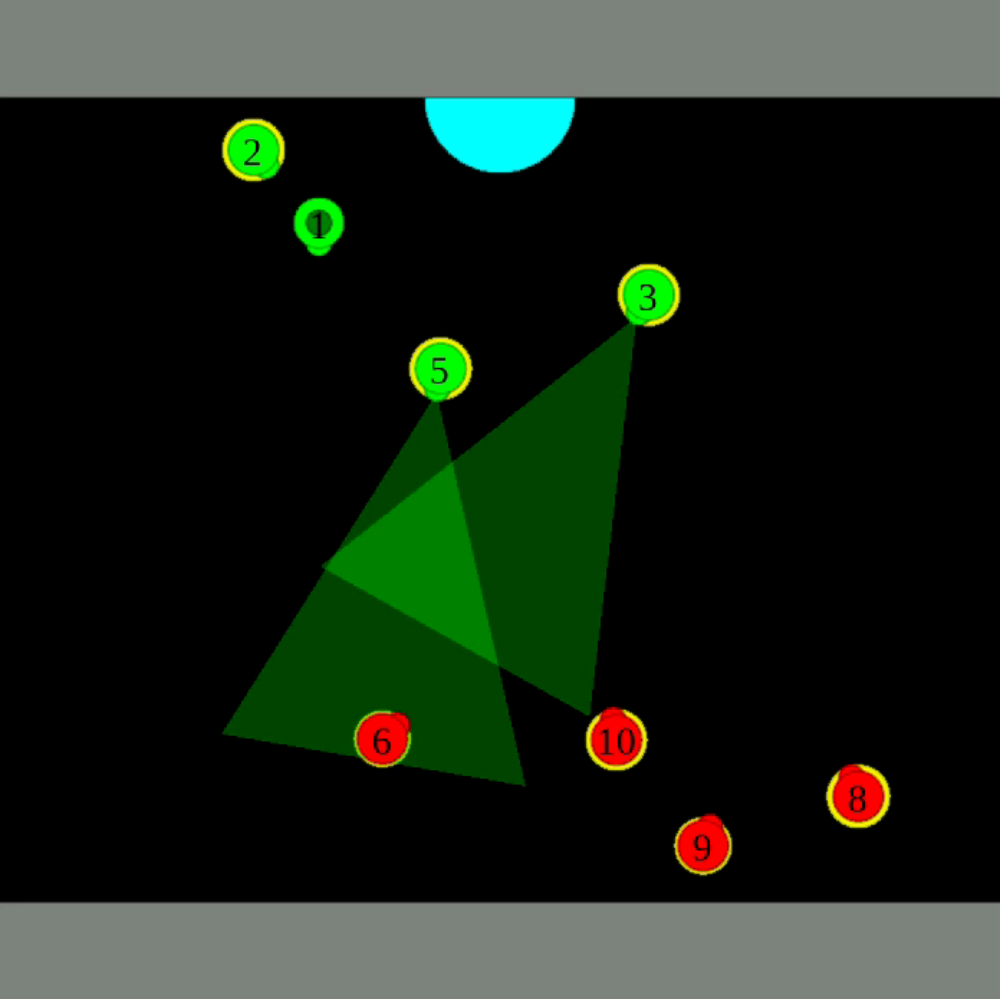}
        \includegraphics[width=0.17\textwidth]{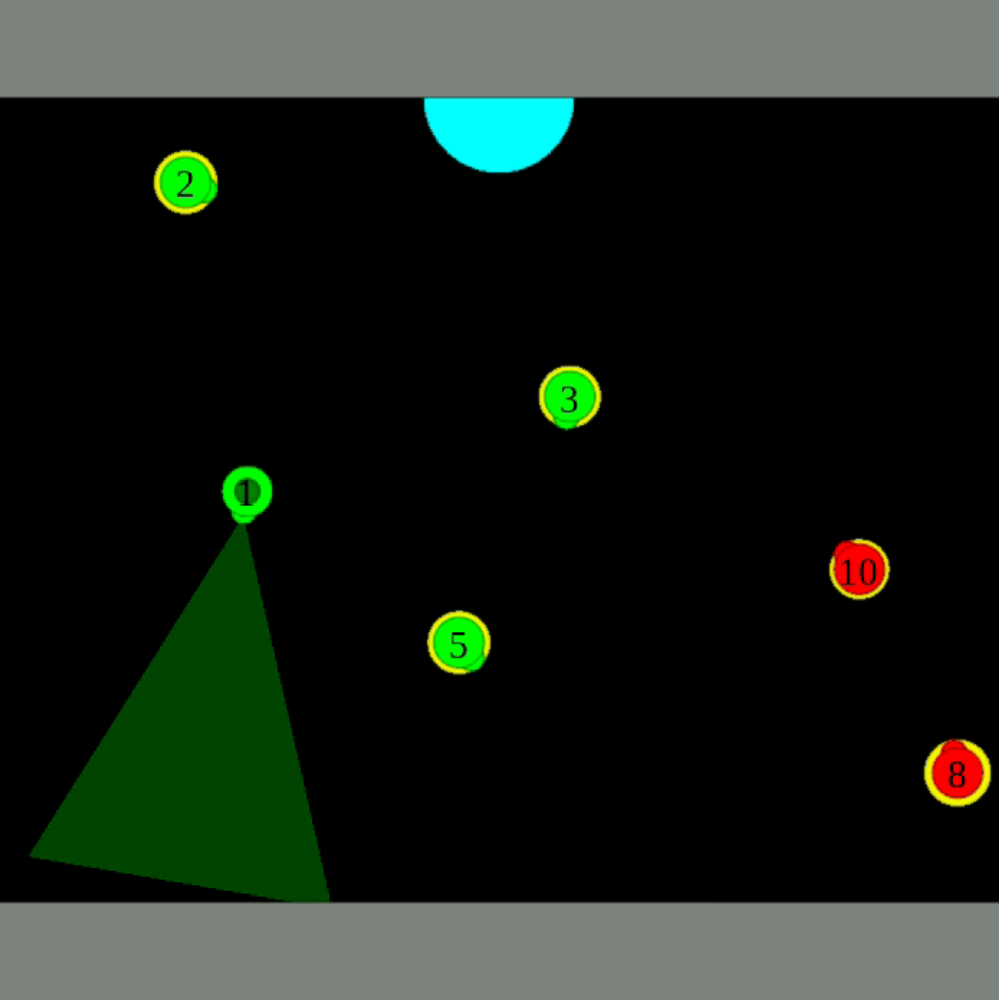}
        \includegraphics[width=0.17\textwidth]{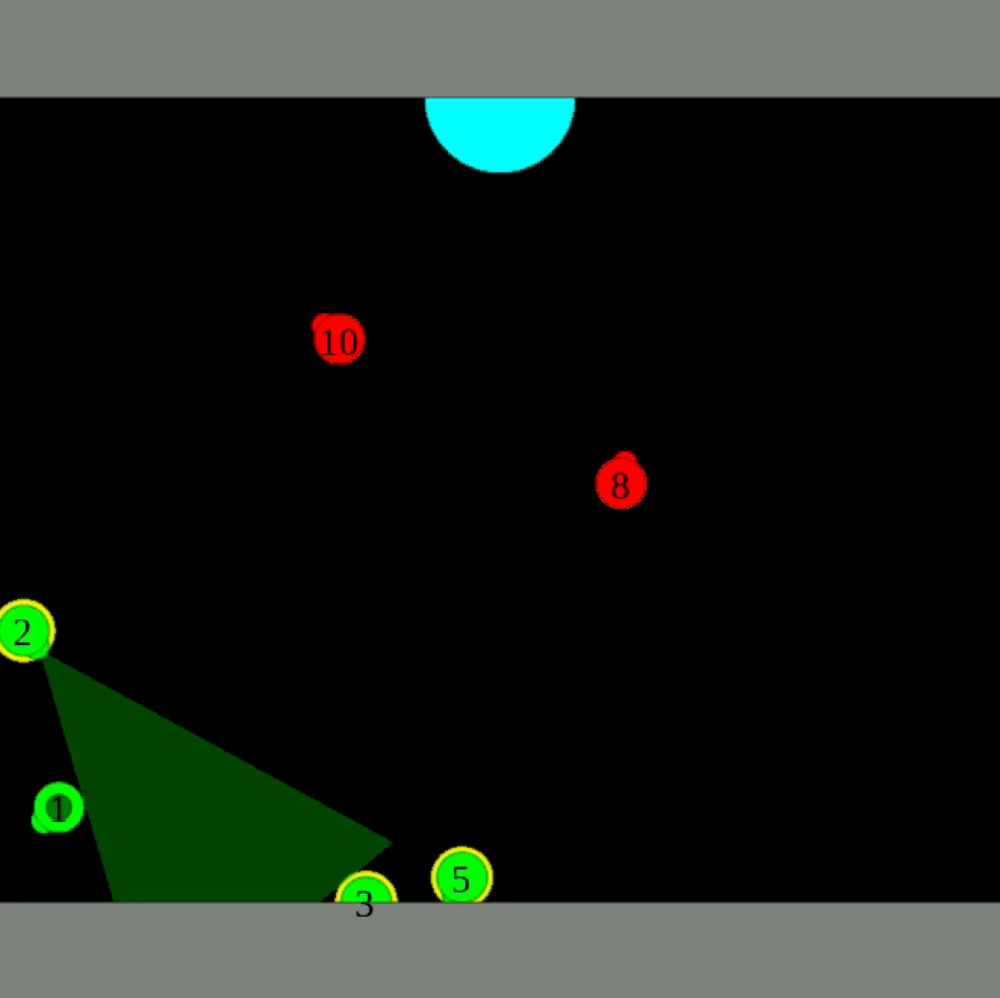}
        
    }
    \vspace{0.1mm}
    \subfloat[Laser flashing strategy of guards]
    {
        \includegraphics[width=0.17\textwidth]{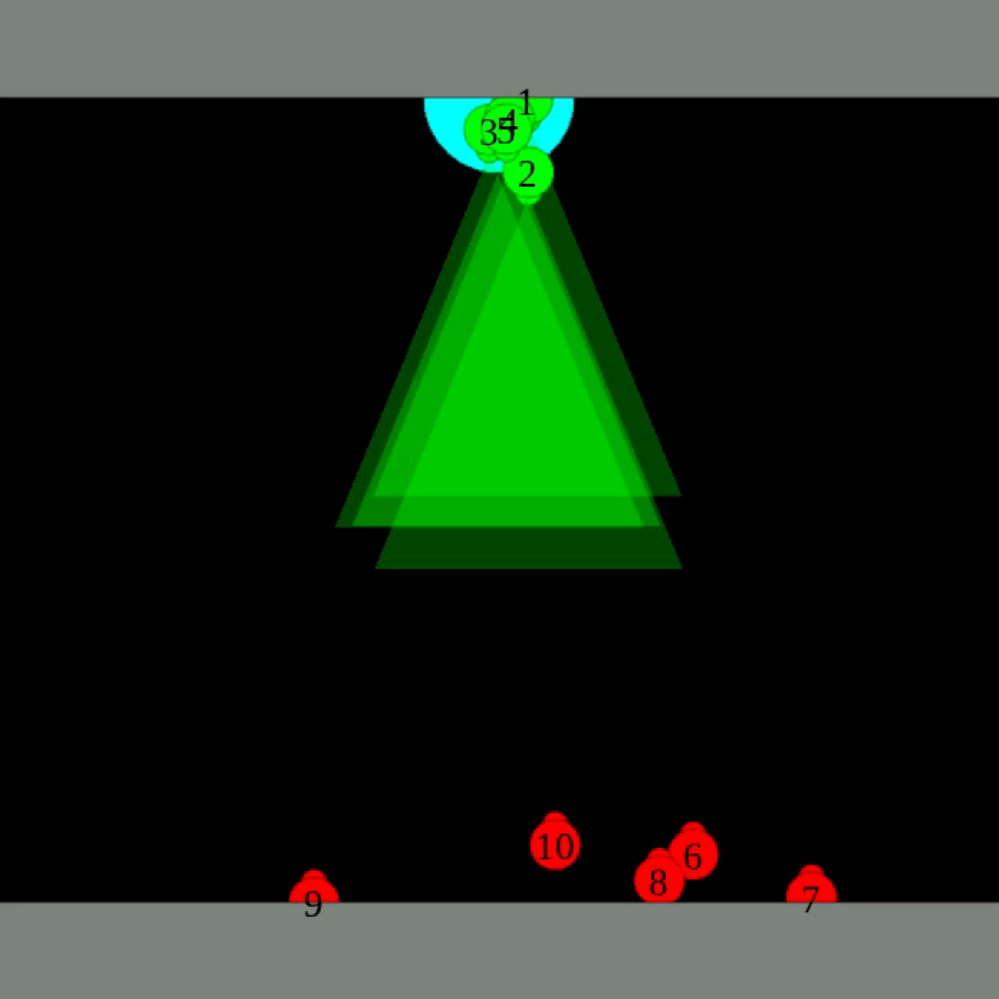}
        \includegraphics[width=0.17\textwidth]{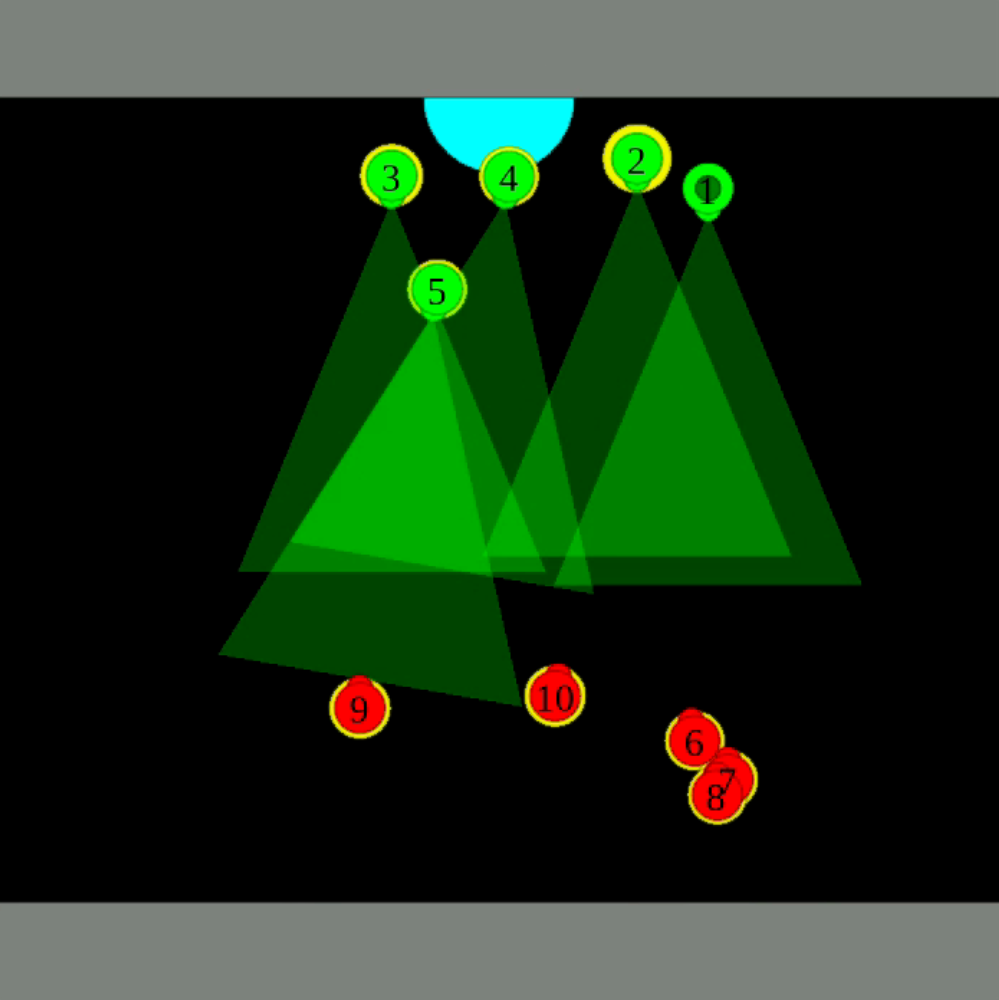}
        \includegraphics[width=0.17\textwidth]{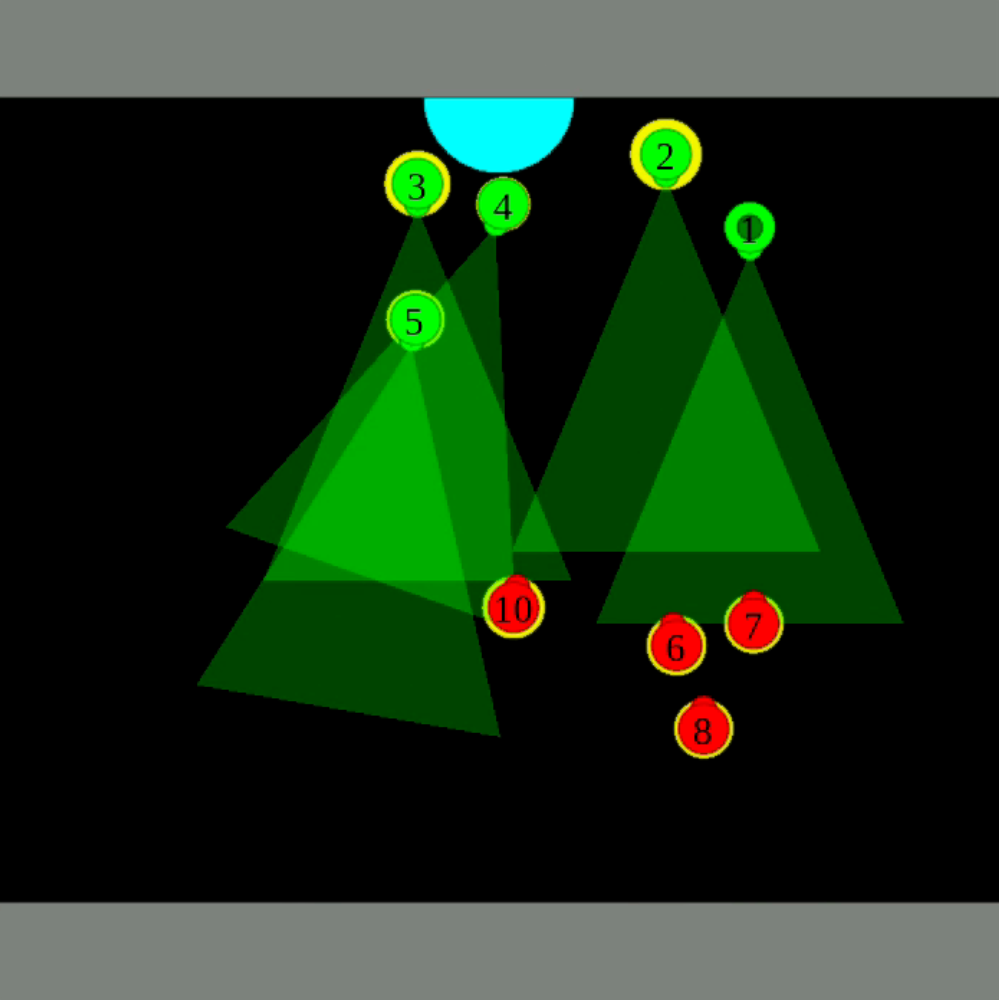}
        \includegraphics[width=0.17\textwidth]{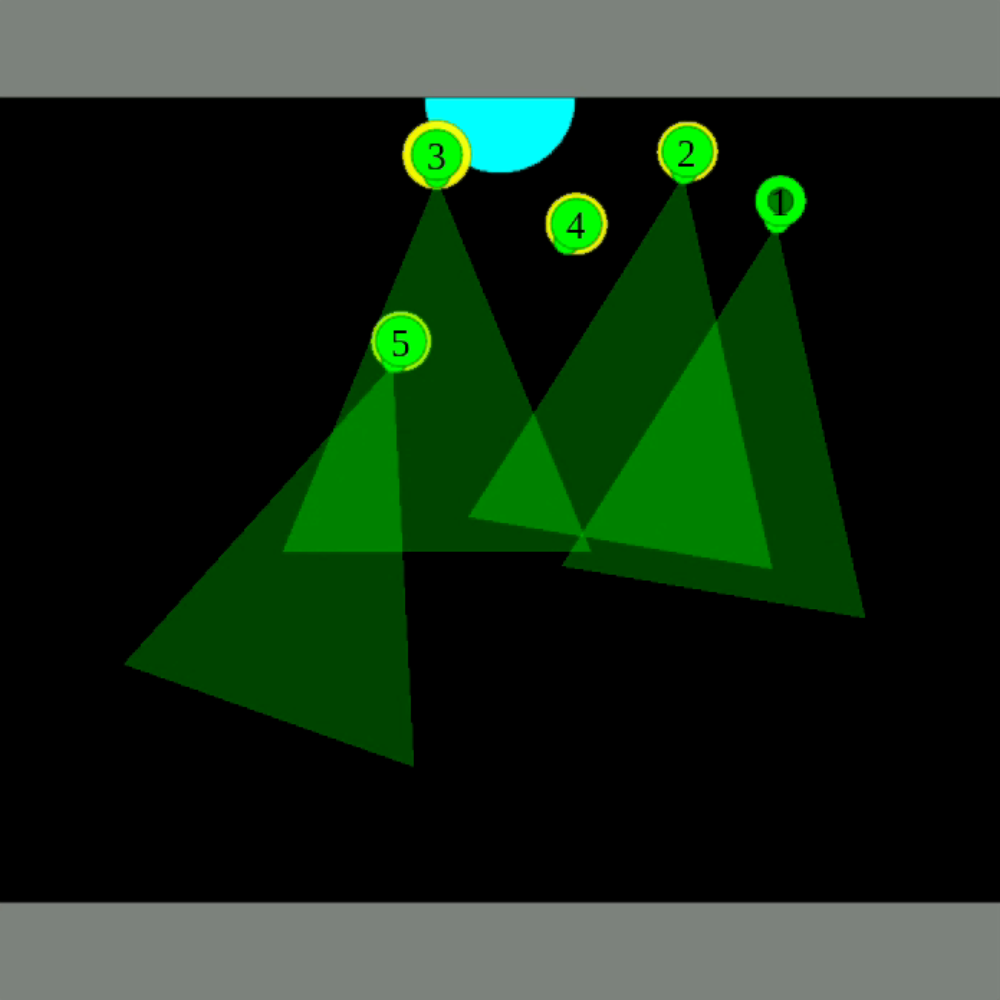}
    }
    \vspace{0.1mm}
    \subfloat[Sneaking strategy of attackers]
    {
        \includegraphics[width=0.17\textwidth]{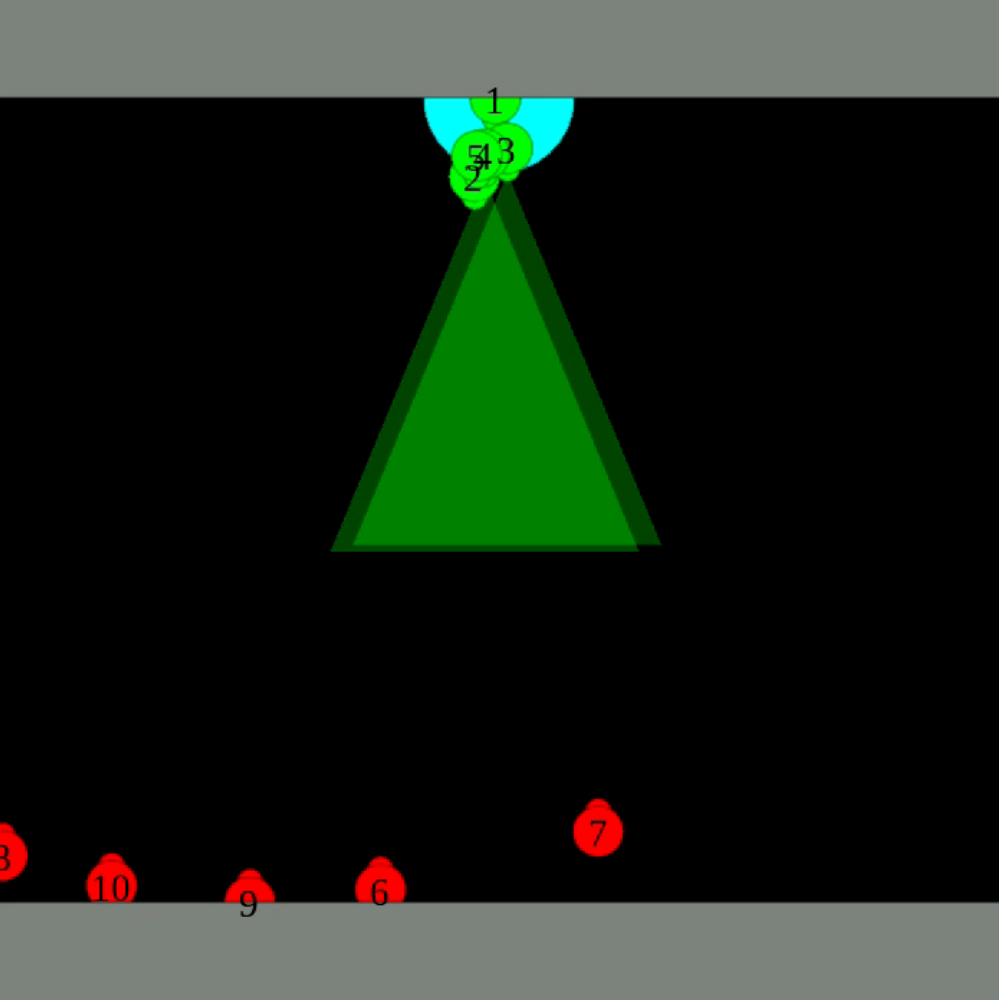}
        \includegraphics[width=0.17\textwidth]{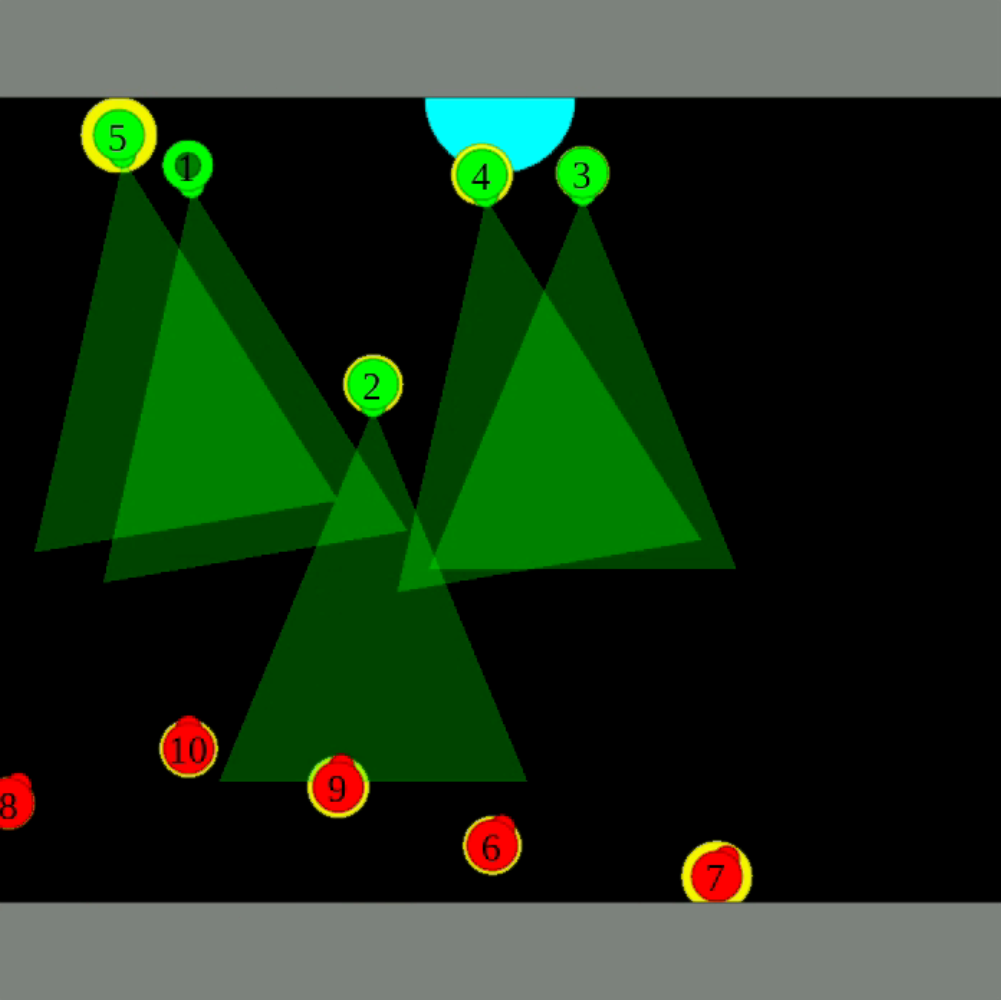}
        \includegraphics[width=0.17\textwidth]{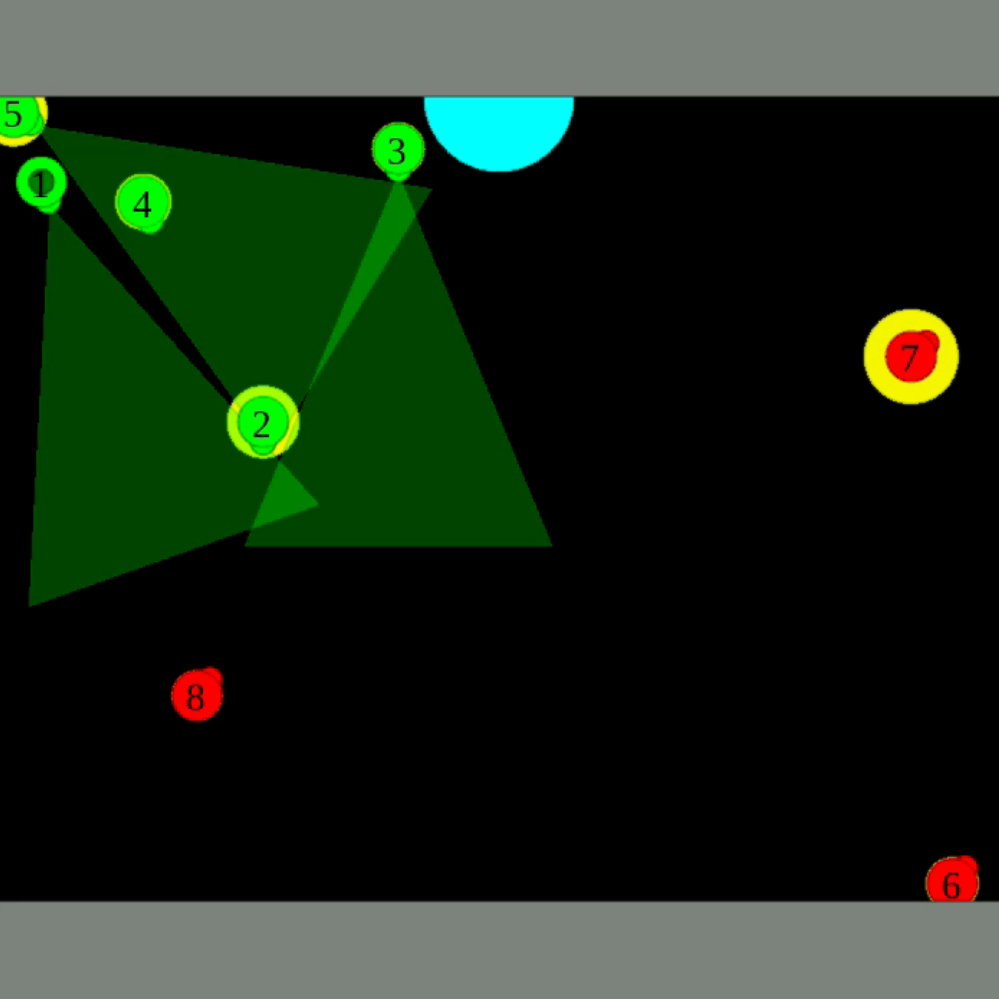}
        \includegraphics[width=0.17\textwidth]{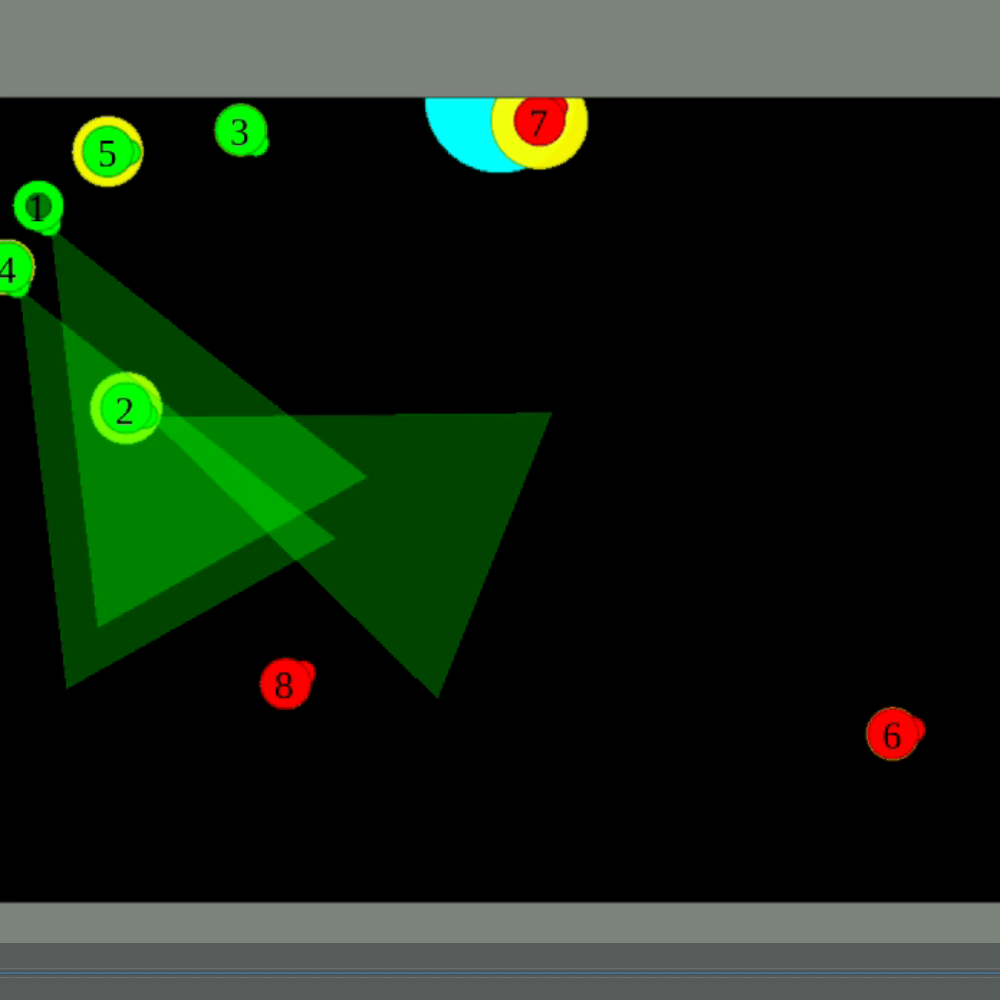}
    }
    \vspace{0.1mm}
    \subfloat[Spreading and flashing strategy of guards]
    {
        \includegraphics[width=0.17\textwidth]{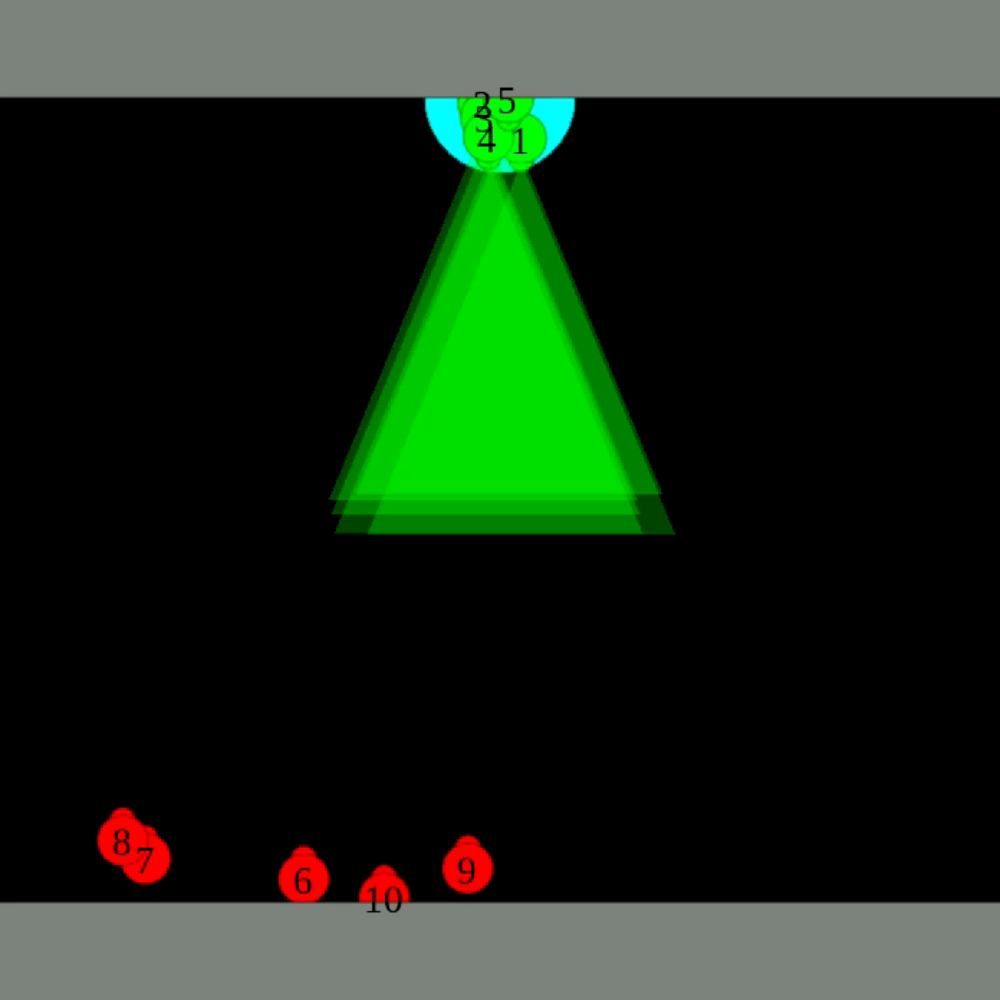}
        \includegraphics[width=0.17\textwidth]{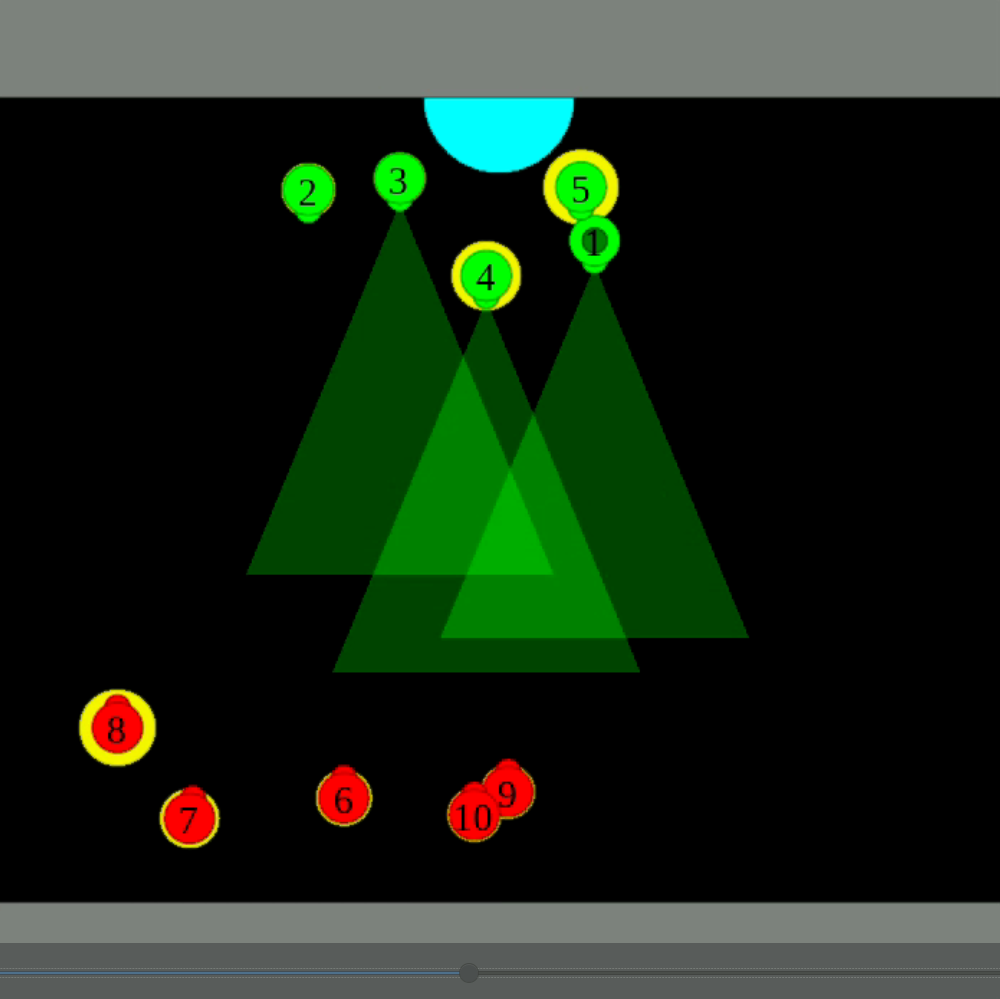}
        \includegraphics[width=0.17\textwidth]{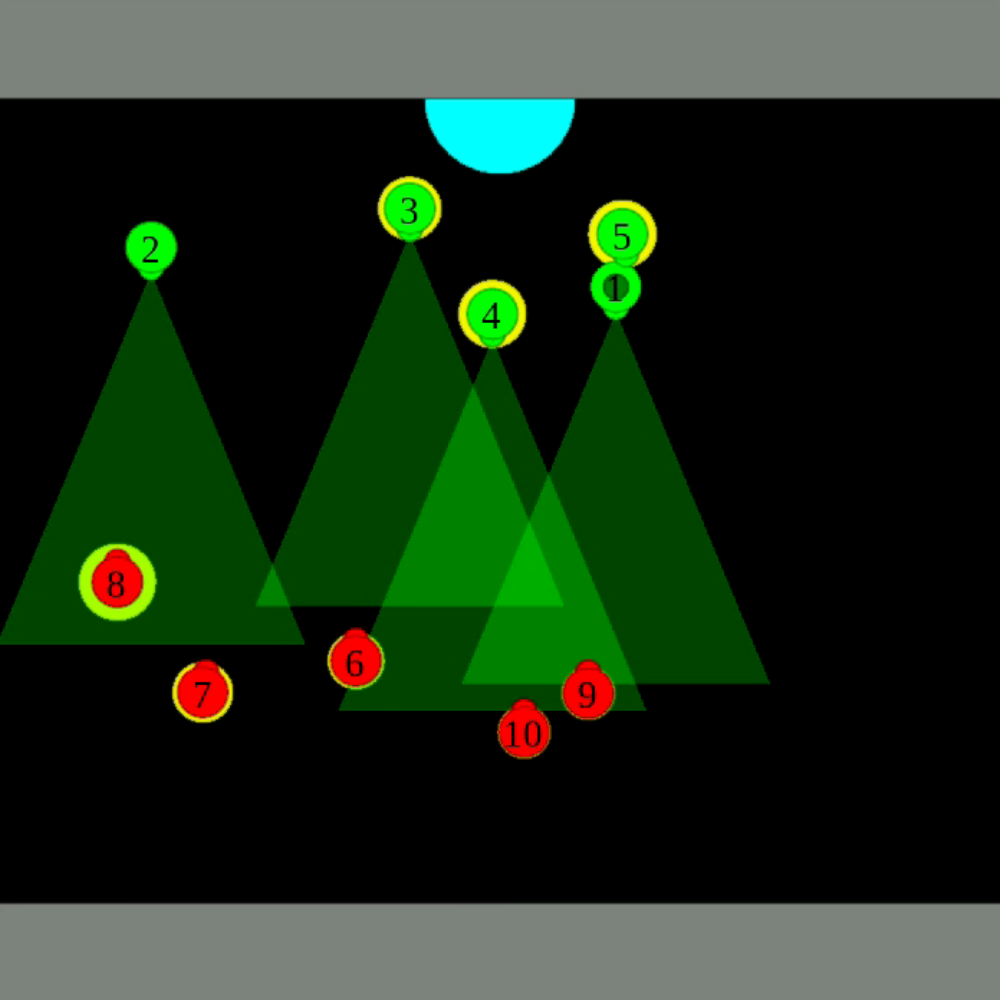}
        \includegraphics[width=0.17\textwidth]{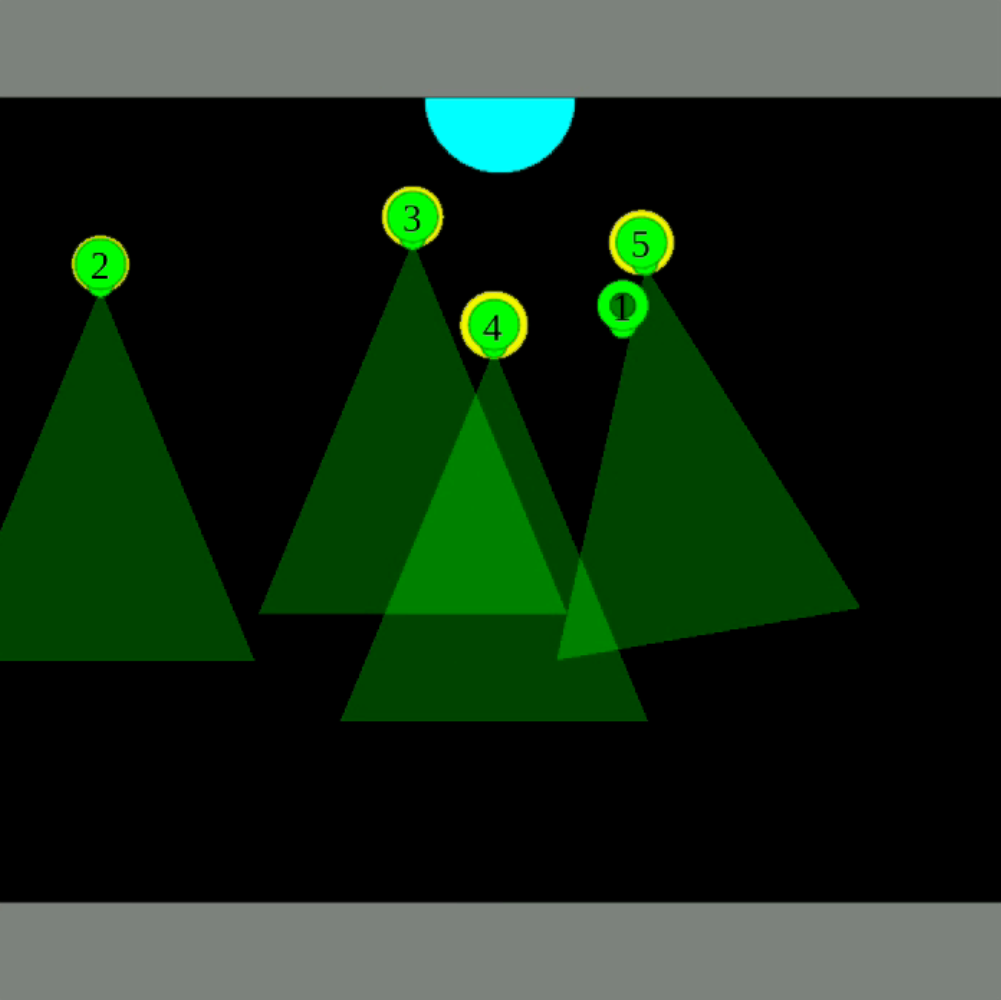}
    }
    \vspace{0.1mm}
    \subfloat[Deception strategy of attackers]
    {
        \includegraphics[width=0.17\textwidth]{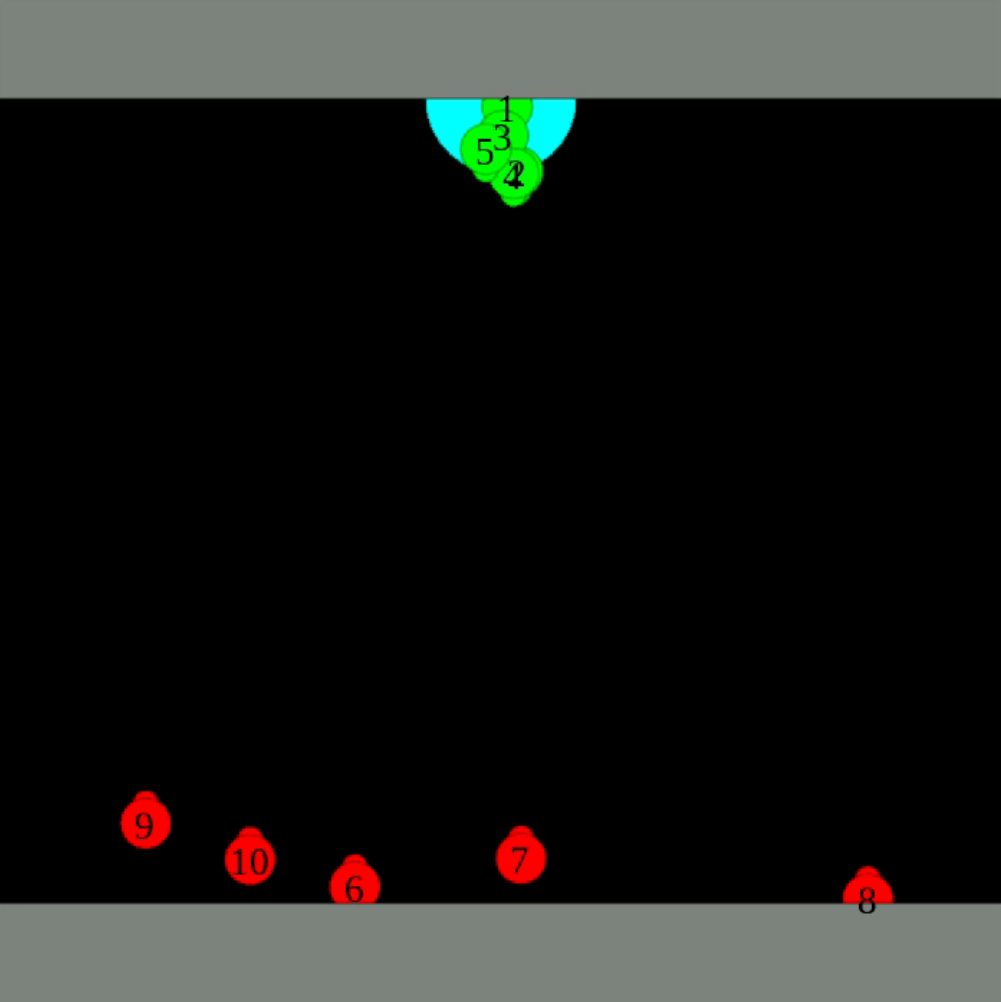}
        \includegraphics[width=0.17\textwidth]{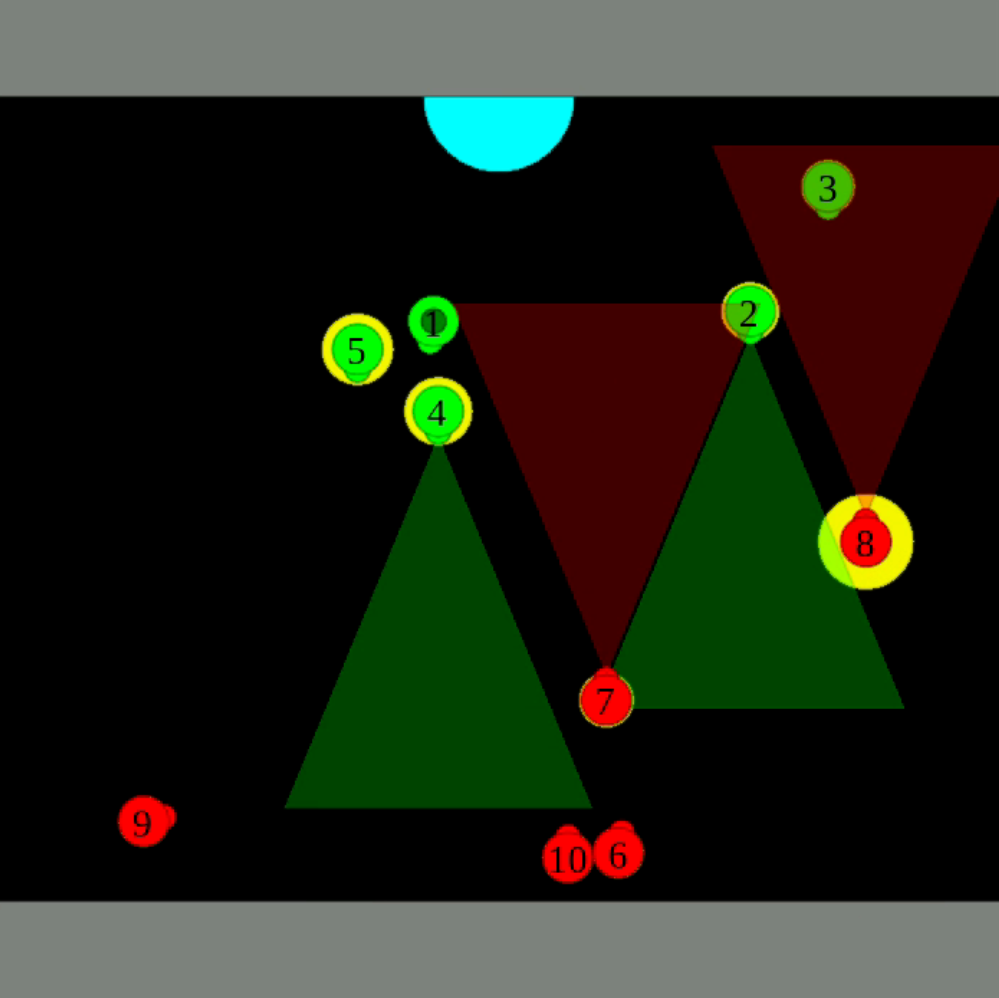}
        \includegraphics[width=0.17\textwidth]{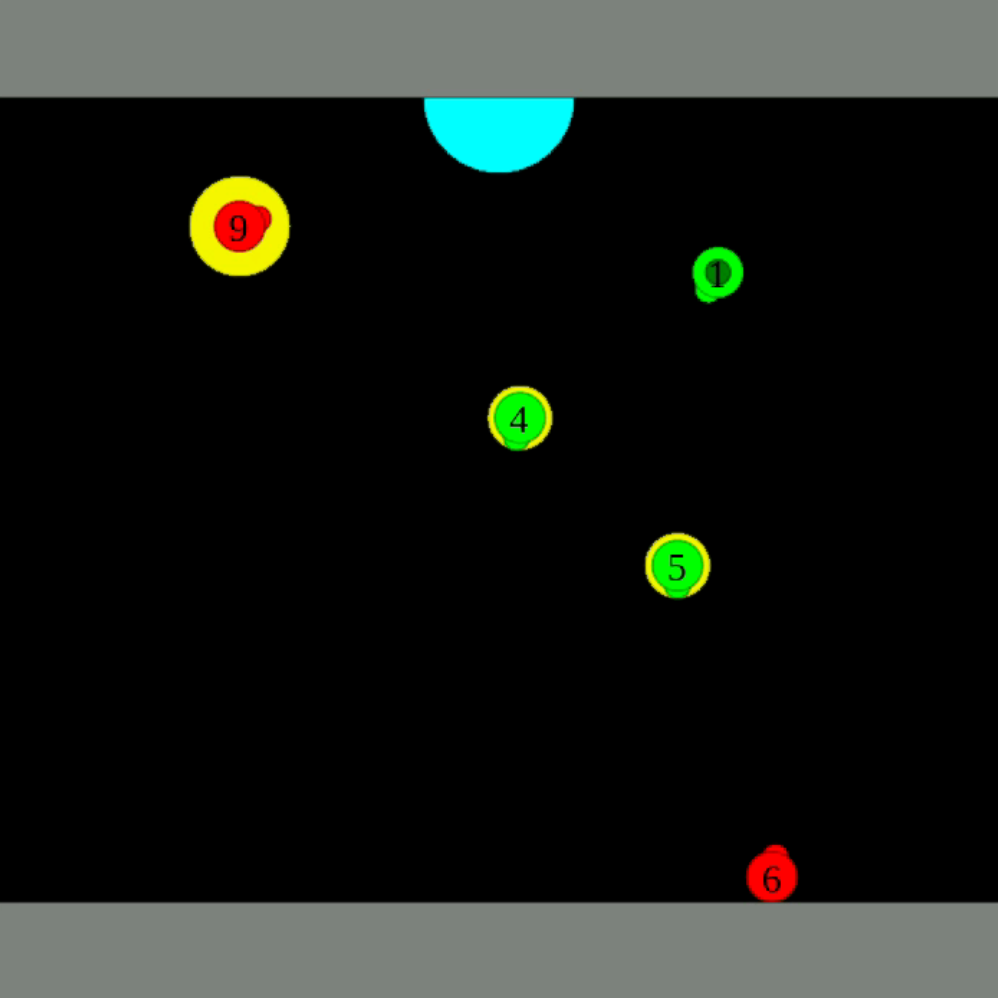}
        \includegraphics[width=0.17\textwidth]{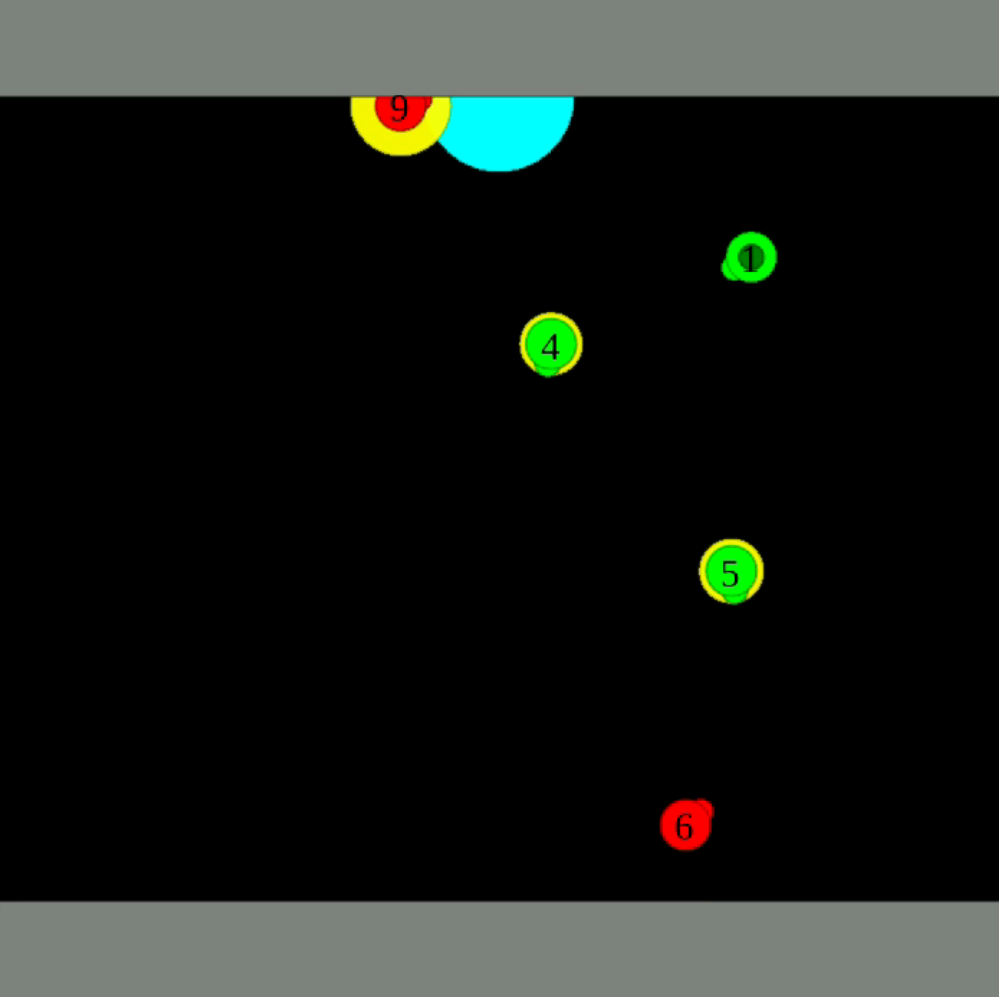}
        \label{fig:seq_deception}
    }
    \vspace{0.1mm}
    \subfloat[Smartly spreading strategy of guards]
    {
        \includegraphics[width=0.17\textwidth]{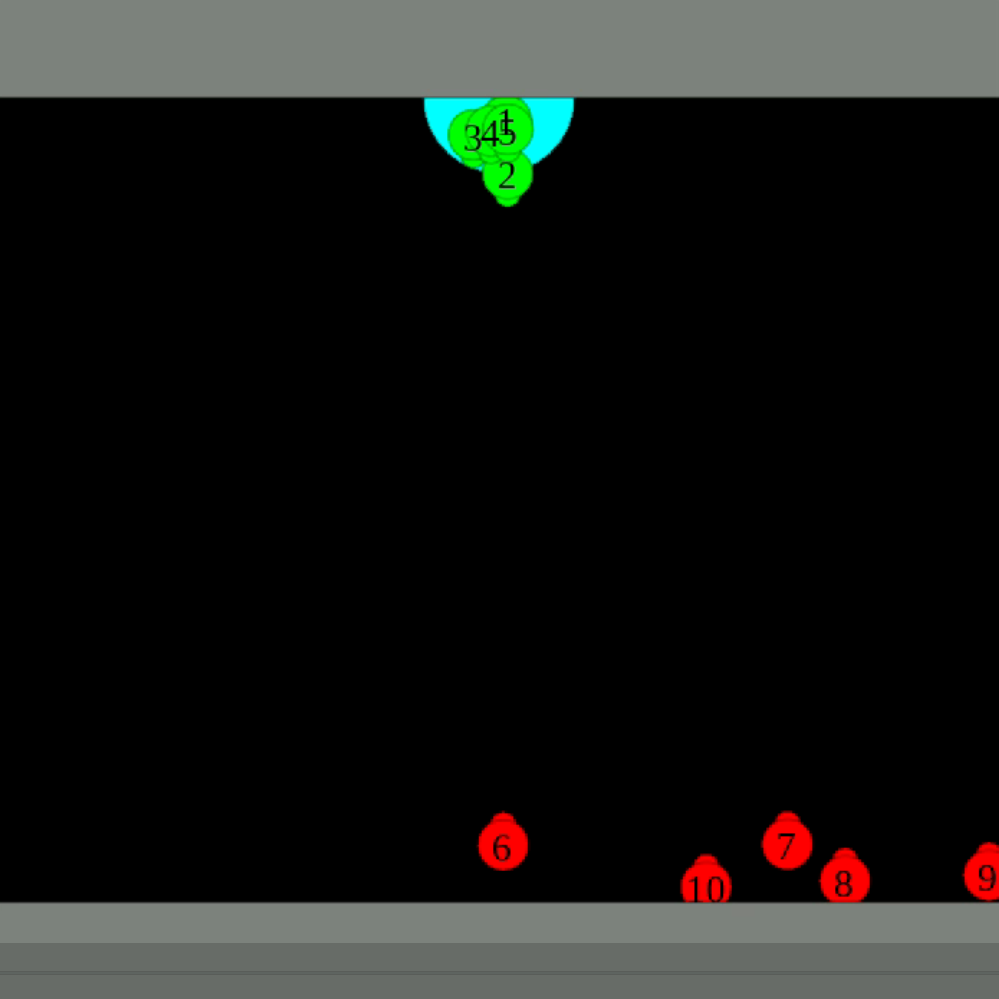}
        \includegraphics[width=0.17\textwidth]{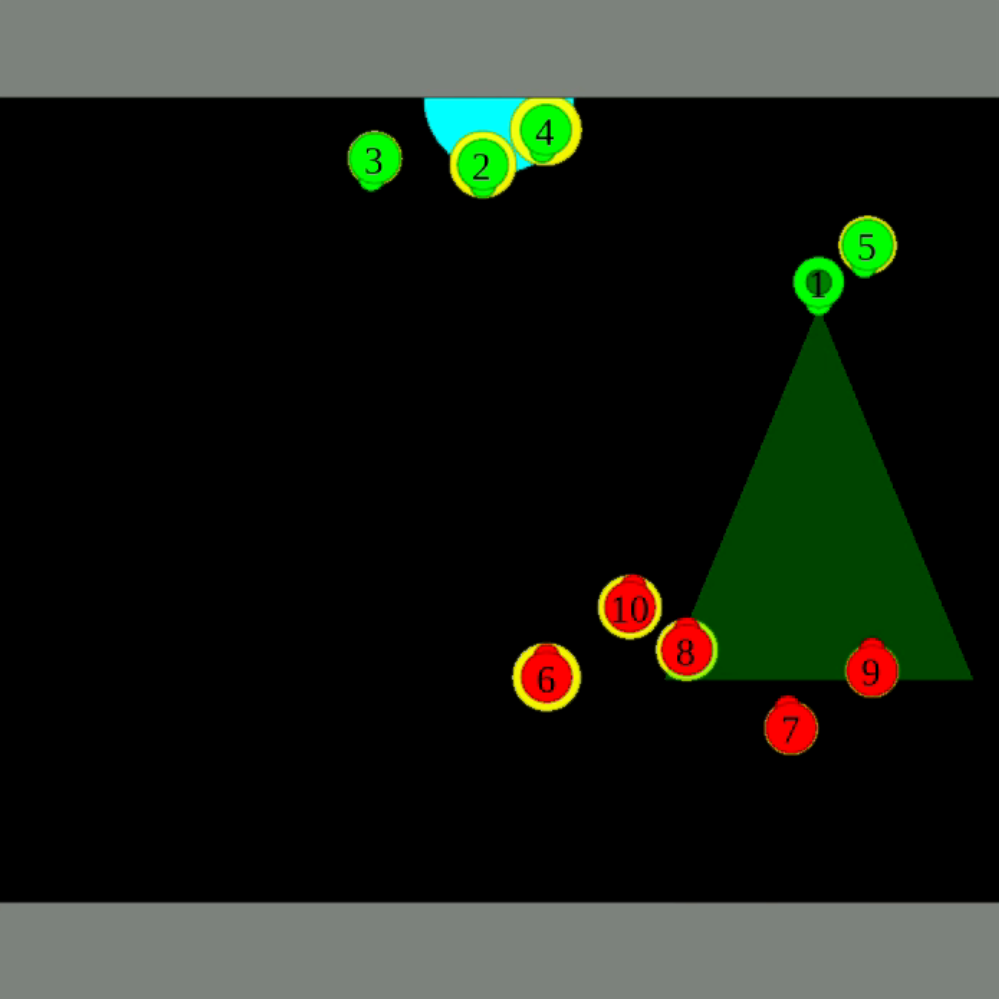}
        \includegraphics[width=0.17\textwidth]{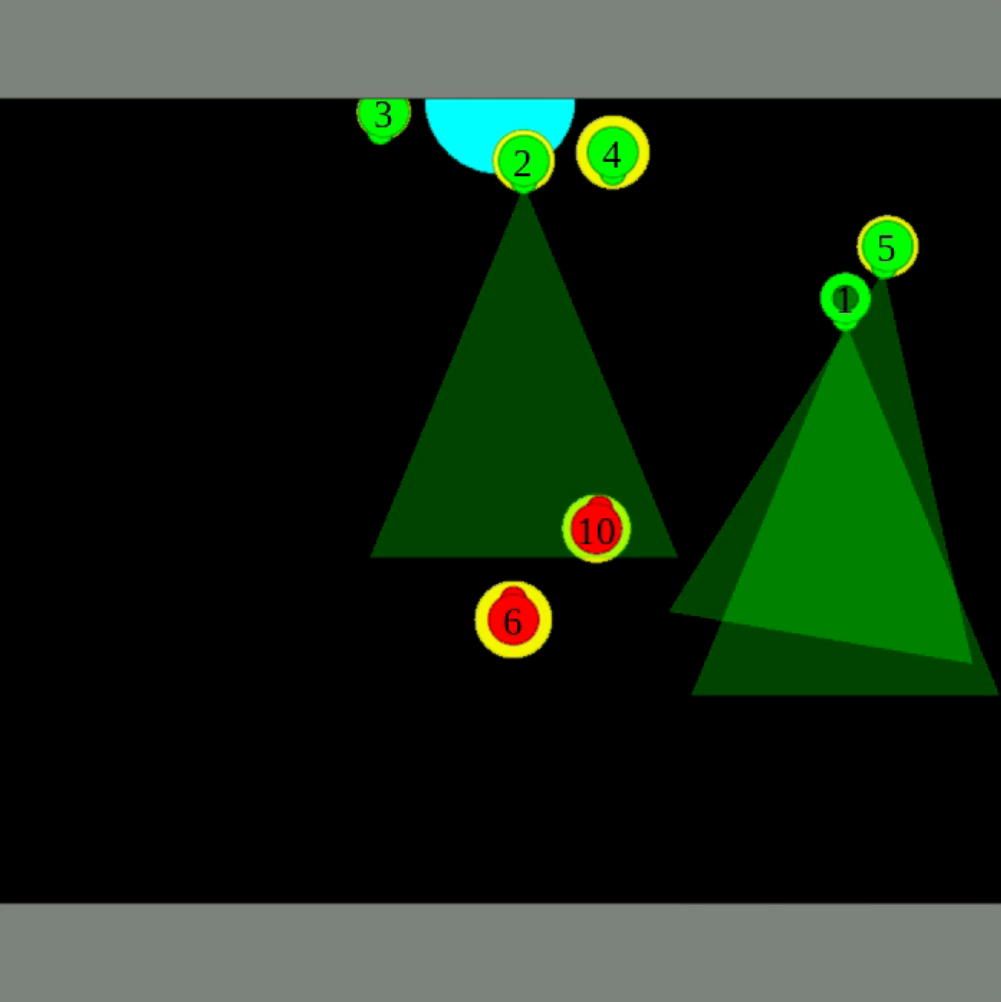}
        \includegraphics[width=0.17\textwidth]{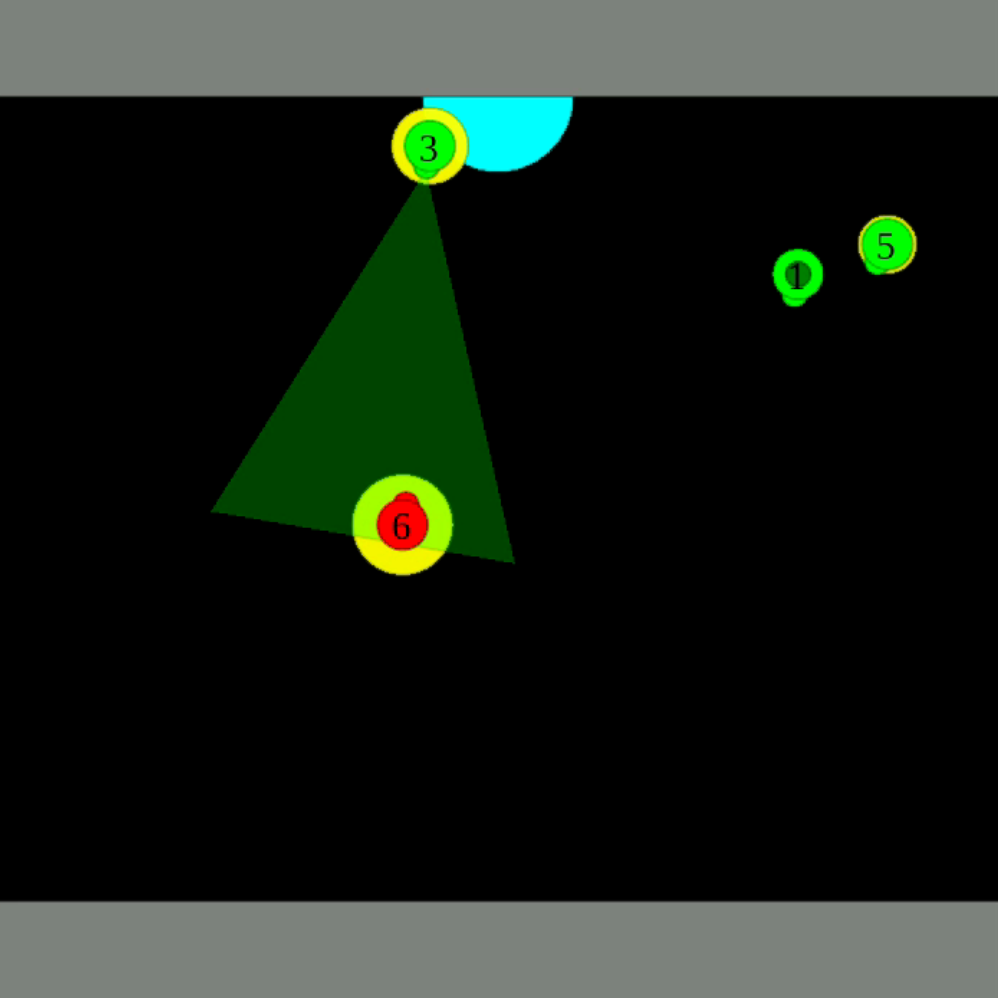}
    }
    \caption{Sample sequences for different strategies that evolved during training. Each row represents one sequence and time moves from left to right.}
    \label{fig:sequences}
\end{figure*}

\section{Results}
We show the results for the 5 guards vs 5 attackers scenario in the FortAttack environment.
\subsection{Evolution of strategies}
Fig. \ref{fig:evolution} shows the reward plot for attackers and guards and snapshots of specific checkpoints as training progresses. The reward for guards is roughly a mirror image of the reward for attackers as victory for one team means defeat for the other. The rewards oscillate with multiple local extrema, i.e. maxima for one team and a corresponding minima for the other. These extrema correspond to increasingly complex strategies that evolve naturally - as one team gets better at its task, it creates pressure for the other team, which in turn comes up with a stronger and more complex strategic behavior.  
\begin{enumerate}[label=(\alph*)]
    \item \emph{Random behavior}: At the beginning of training, agents randomly move around and shoot in the wild. They explore trying to make sense of the FortAttack environment and their goals in this world.
    \item \emph{Flash laser}: Attackers eventually learn to approach the fort and the guards adopt a simple strategy to win. They all continuously flash their lasers creating a protection zone in front of the fort which kills any attacker that tries to enter.
    \item \emph{Sneak}: As guards block entry from the front, attackers play smart. They approach from all the directions, some of them get killed but one of them manages to sneak in from the side.
    \item \emph{Spread and flash}: In response to the sneaking behavior, the guards learn to spread out and kill all attackers before they can sneak in.
    \item \emph{Deceive}: To tackle the strong guards, the attackers come up with the strategy of deception. Most of them move forward from the right while one holds back on the left. The guards start shooting at the attackers on the right which diverts their attention from the single attacker on the left. This attacker quietly waits for the right moment to sneak in, bringing victory for the whole team. Note that this strategy requires heterogeneous behavior amongst the homogeneous agents, which naturally evolved without explicitly being encouraged to do so.
    \item \emph{Spread smartly}: In response to this, the guards learn to spread smartly, covering a wider region and killing attackers before they can sneak in.
    
\end{enumerate}

\subsection{Being Attentive}
In each of the environment snapshots in Fig. \ref{fig:evolution} and Fig. \ref{fig:sequences}, we visualize the attention paid by one alive guard to all the other agents. This guard has a dark green dot at it's center. All the other agents have yellow rings around them, with the sizes of the rings being proportional to the attention values. Eg. in Fig. \ref{fig:sequences}(e), agent 1 initially paid roughly uniform and low attention to all attackers when they were far away. Then, it started paying more attention to agent 8, which was attacking aggressively from the right. Little did it know that it was being deceived by the clever attackers. When agent 9 reached near the fort, agent 1 finally started paying more attention to the sneaky agent 9 but it was too late and the attackers had successfully deceived it.

\subsection{Ensemble strategies}
\begin{figure}[h!]
    \centering
    \includegraphics[width=\columnwidth]{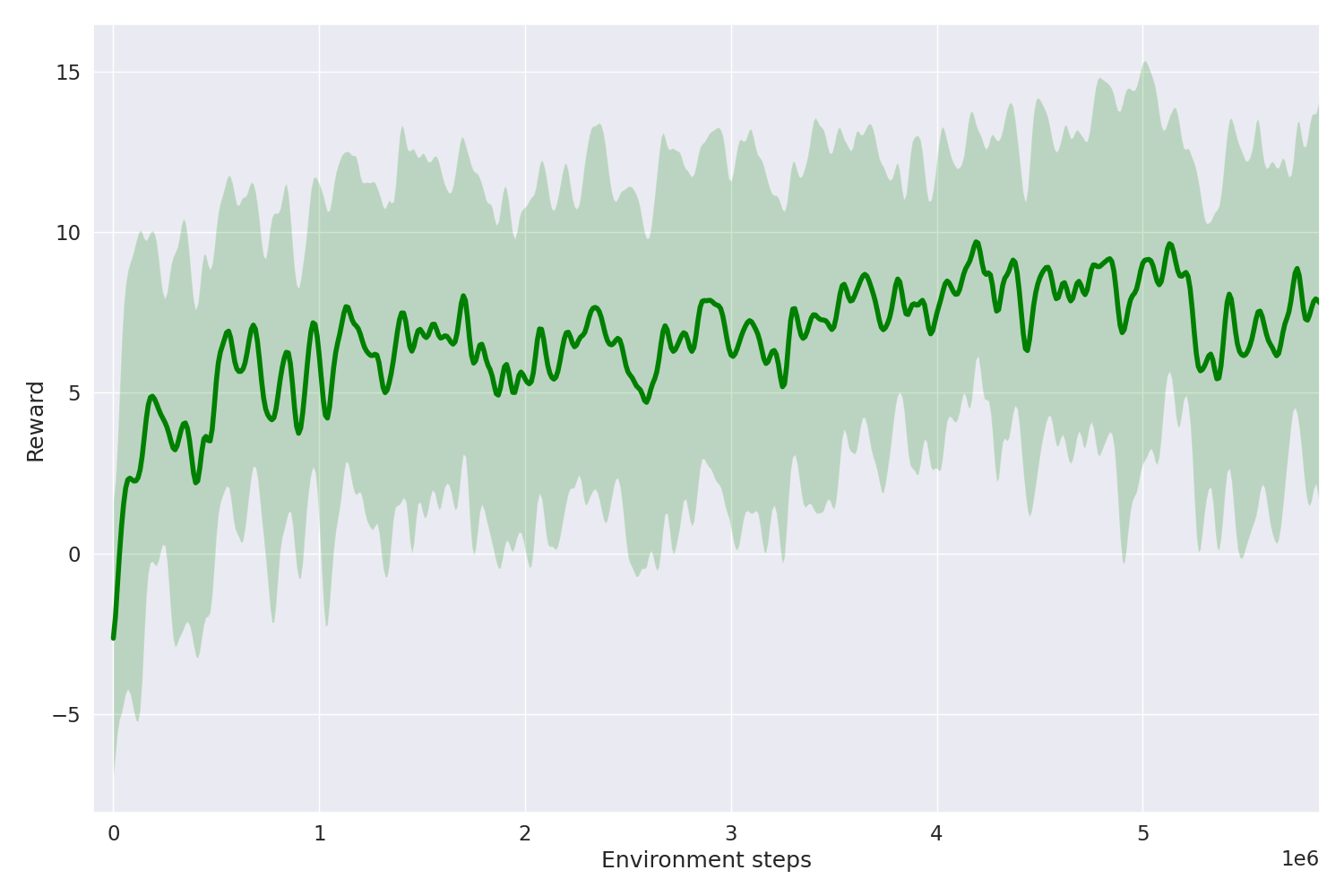}
    \caption{Average reward per agent per episode for guards as ensemble training progresses. The reward is shown after Gaussian smoothing.}
    \label{fig:ensemble_training}
\end{figure}

To train and generate strong agents, we first need strong opponents to train against. The learnt strategies in the previous section give us a natural way to generate strategies from simple rules of the game. If we wish to get strong guards, we can train a single guard policy against all of the attacker strategies, by randomly sampling one attacker strategy for each environment episode. Fig. \ref{fig:ensemble_training} shows the reward for guards as training progresses. This time, the reward for guards continually increases and doesn't show an oscillating behavior.

\section{Conclusions}
In this work we were able to scale to multiple agents by modeling inter agent interactions with a graph containing two attention layers. We studied the evolution of complex multi agent strategies in a mixed cooperative-competitive environment. In particular, we saw the natural emergence of deception strategy which required heterogeneous behavior amongst homogeneous agents. If instead we wanted to explicitly encode heterogeneous strategies, a simple extension of our work would be to have different sets of policy parameters ($f_{\theta_d}$) within the same team, eg. one set for aggressive guards and one set of defensive guards.
We believe that our study would inspire further work towards scaling multi agent reinforcement learning to large number of agents in more complex mixed cooperative-competitive scenarios.




\section*{Acknowledgment}
{We would like to thank the authors of \cite{agarwal2019learning} and \cite{lowe2017multi} for open sourcing their code repositories which helped us in our implementation. We would also like to thank Advait Gadhikar, Tejus Gupta, Wenhao Luo and Yash Belhe for their insightful comments.}

\bibliographystyle{IEEEtran}
\bibliography{Fortattack}

\clearpage
\section*{Appendix}
\subsection{Reward structure}
\begin{table}[h!]
    \centering
    \caption{Reward structure}
    \newcounter{sl}
    \begin{tabular}{|p{0.05\columnwidth}|p{0.35\columnwidth}|p{0.45\columnwidth}|}
         \hline
         \textbf{Sl. No.} & \textbf{Event} & \textbf{Reward} \\
         \hline
         \stepcounter{sl}
         \arabic{sl} & Guard $i$ leaves the fort & Guard $i$ gets -ve reward.\\
         \hline
         \stepcounter{sl}
         \arabic{sl} & Guard $i$ returns to the fort & Guard $i$ gets +ve reward.\\
         \hline
         \stepcounter{sl}
         \arabic{sl} & Attacker $j$ moves closer to the fort & Attacker $j$ gets small +ve reward\\
         \hline
         \stepcounter{sl}
         \arabic{sl} & Attacker $j$ moves away from the fort & Attacker $j$ gets small -ve reward\\
         \hline
         \stepcounter{sl}
         \arabic{sl} & Guard $i$ shoots attacker $j$ with laser & Guard $i$ gets +ve reward and attacker $j$ gets -ve reward.\\
         \hline
         \stepcounter{sl}
         \arabic{sl} & Attacker $j$ shoots guard $i$ with laser & Guard $i$ gets -ve reward and attacker $j$ gets +ve reward.\\
         \hline
         \stepcounter{sl}
         \arabic{sl} & Agent $i$ shoots laser but doesn't hit any opponent & Agent $i$ gets low -ve reward.\\
         \hline
         \stepcounter{sl}
         \arabic{sl} & All attackers are killed & All alive guards get high +ve reward. Attacker(s) that just got killed gets high -ve reward.\\
         \hline
         \stepcounter{sl}
         \arabic{sl} & Attacker $j$ reaches the fort & All alive guards high -ve reward. Attacker $j$ gets high +ve reward.\\
         \hline
    \end{tabular}
    \label{tab:reward_structure}
\end{table}

Table \ref{tab:reward_structure} describes the reward structure for the FortAttack environment.
The negative reward for wasting a laser shot is higher in magnitude for attackers than for guards. Otherwise, we observed that the attackers always managed to win. This reward structure can also be attributed to the fact that attackers in a real world scenario would like to sneak in and wouldn't want to shoot too often and reveal themselves to the guards.

\end{document}